\documentclass[10pt,twocolumn,letterpaper]{article}

\usepackage[pagenumbers]{cvpr} % To force page numbers, e.g. for an arXiv version

\usepackage{times}
\usepackage{epsfig}
\usepackage{graphicx}
\usepackage{amsmath}
\usepackage{ifthen}
\usepackage{amssymb}\usepackage{cite}
\usepackage{subcaption}
\usepackage{multicol}
\usepackage{multirow}
\usepackage{algorithm}
\usepackage{algpseudocode}
\usepackage{bm}
\usepackage{makecell}
\usepackage{arydshln}
\usepackage{pifont}
\usepackage{booktabs}
\usepackage{mathtools}
\usepackage{adjustbox}
\usepackage{wrapfig}

\usepackage[]{xcolor}
\definecolor{cvprblue}{rgb}{0.21,0.49,0.74}
\usepackage[pagebackref,breaklinks,colorlinks,citecolor=cvprblue]{hyperref}

\newcommand{\rf}[1]{{\textbf{\color{red}{#1}}}} % green bold for table entries
\newcommand{\bd}[1]{{\color{blue}{\underline{#1}}}} % blue underlined for table entries
\newcommand{\method}[1]{SUPIR}
\newcommand\blfootnote[1]{%
  \begingroup
\renewcommand\thefootnote{}\footnote{#1}%
  \addtocounter{footnote}{-1}%
  \endgroup
}

\def\paperID{2825} % *** Enter the Paper ID here
\def\confName{CVPR}
\def\confYear{2024}

\title{Scaling Up to Excellence:\\Practicing Model Scaling for Photo-Realistic Image Restoration In the Wild}

\author{Fanghua Yu$^{1,*}$, Jinjin Gu$^{2,*}$, Zheyuan Li$^{1}$, Jinfan Hu$^{1}$, Xiangtao Kong$^{3}$,\\Xintao Wang$^{4}$, Jingwen He$^{2,5}$, Yu Qiao$^{2}$, Chao Dong$^{1,2,\dag}$\\
\small{$^1$Shenzhen Institute of Advanced Technology, Chinese Academy of Sciences}\quad
\small{$^2$Shanghai AI Laboratory} \\
\small{$^3$The Hong Kong Polytechnic University}\quad
\small{$^4$ARC Lab, Tencent PCG}\quad
\small{$^5$The Chinese University of Hong Kong}
\\
{\tt\small Project Page: \url{https://supir.xpixel.group}}
}

\begin{document}

\setlength{\abovedisplayskip}{3pt}
\setlength{\belowdisplayskip}{2pt}

\twocolumn[{
	\renewcommand\twocolumn[1][]{#1}
	\maketitle
	\thispagestyle{empty}
	\vspace{-30pt}
	\begin{center}
		\includegraphics[width=\linewidth]{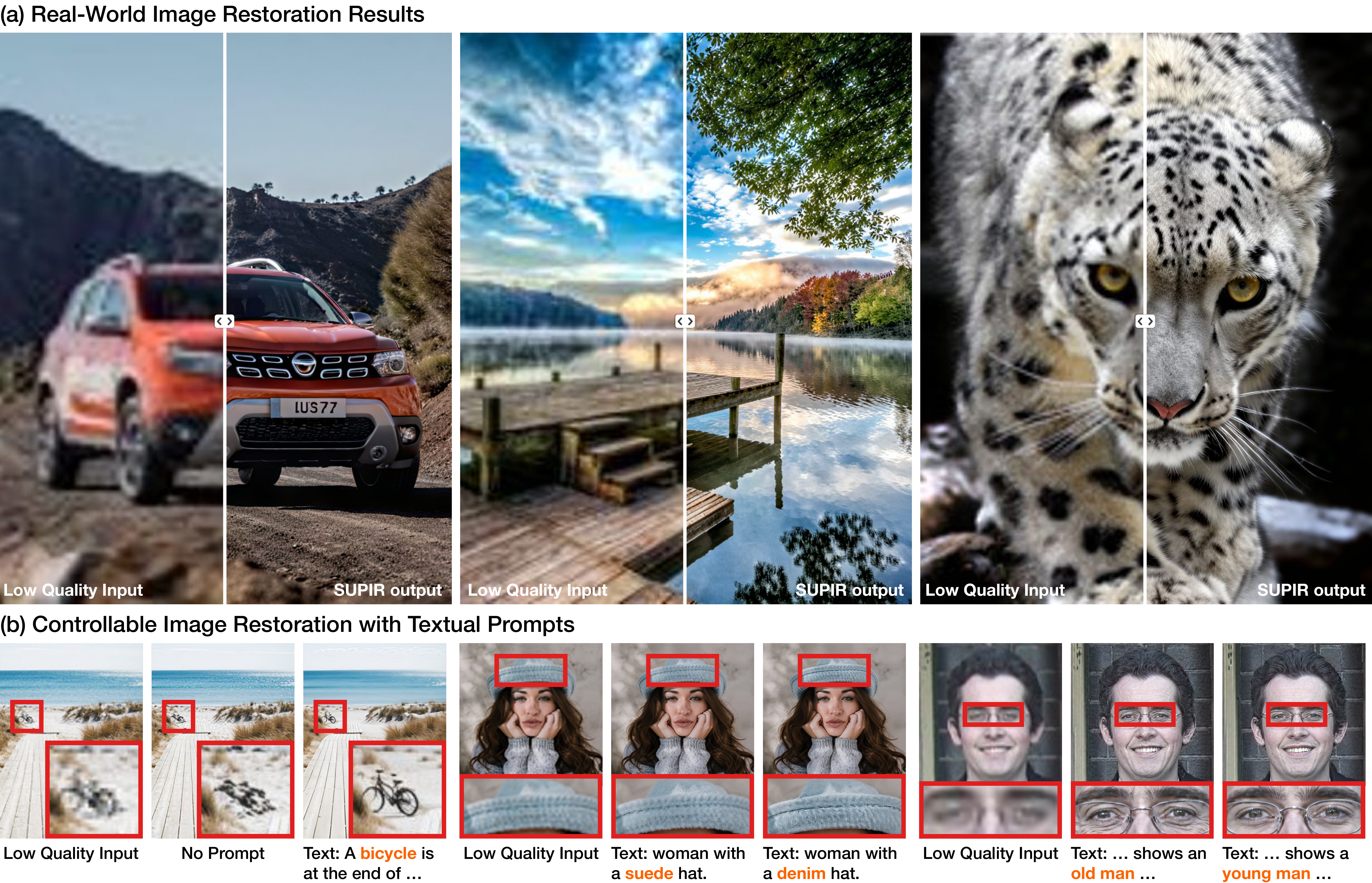}
		\vspace{-18pt}
            \captionof{figure}{
                Our SUPIR model demonstrates remarkable restoration effects on real-world low-quality images, as illustrated in (a). Additionally, SUPIR features targeted restoration capability driven by textual prompts. For instance, it can specify the restoration of blurry objects in the distance (case 1), define the material texture of objects (case 2), and adjust restoration based on high-level semantics (case 3).
            }
		\label{fig:teaser}
		\vspace{5pt}
	\end{center}
}]

\maketitle
\begin{abstract}
\vspace{-10pt}
\blfootnote{
\noindent $^*$ Contribute Equally. \quad $^\dag$ Corresponding Author.
}
We introduce SUPIR (Scaling-UP Image Restoration), a groundbreaking image restoration method that harnesses generative prior and the power of model scaling up.
Leveraging multi-modal techniques and advanced generative prior, SUPIR marks a significant advance in intelligent and realistic image restoration.
As a pivotal catalyst within SUPIR, model scaling dramatically enhances its capabilities and demonstrates new potential for image restoration.
We collect a dataset comprising 20 million high-resolution, high-quality images for model training, each enriched with descriptive text annotations.
SUPIR provides the capability to restore images guided by textual prompts, broadening its application scope and potential. 
Moreover, we introduce negative-quality prompts to further improve perceptual quality.
We also develop a restoration-guided sampling method to suppress the fidelity issue encountered in generative-based restoration.
Experiments demonstrate SUPIR's exceptional restoration effects and its novel capacity to manipulate restoration through textual prompts.
\end{abstract}    
% \clearpage
\vspace{-4mm}
\section{Introduction}
\label{sec:intro}
\vspace{-2mm}
The development of image restoration (IR) has greatly elevated expectations for both the perceptual effects and the intelligence of IR results.
IR methods based on generative priors \cite{kawar2022denoising,wang2022zero,rombach2022high,lin2023diffbir} leverage powerful pre-trained generative models to introduce high-quality generation and prior knowledge into IR, bringing significant progress in these aspects.
Continuously improving the capabilities of the generative prior is key to achieving better IR results, with model scaling being a crucial and effective approach.
There are many tasks that have obtained astonishing improvements from scaling, such as SAM \cite{kirillov2023segment} and large language models (LLMs) \cite{brown2020language,touvron2023llama,2023internlm}.
This further drives our pursuit of constructing large-scale, intelligent IR models that can produce ultra-high-quality images.
However, due to engineering constraints such as computing resources, model architecture, training data, and the cooperation of generative models and IR, scaling up IR models is challenging.

In this work, we introduce SUPIR (Scaling-UP IR), the largest-ever IR method, aimed at exploring greater potential in restoration visual effects and intelligence.
Specifically, SUPIR employs StableDiffusion-XL (SDXL) \cite{podell2023sdxl} as a powerful generative prior, which contains 2.6 billion parameters.
To effectively deploy this model in IR, we design and train a large-scale adaptor that incorporates a novel component named the ZeroSFT connector.
To maximize the benefits of model scaling, we collect a dataset of over 20 million high-quality, high-resolution images, each accompanied by detailed descriptive text.
We utilize a 13-billion-parameter multi-modal language model to provide image content prompts, greatly improving the accuracy and intelligence of our method.
The proposed SUPIR model demonstrates exceptional performance in a variety of IR tasks, achieving the best visual quality, especially in complex and challenging real-world scenarios.
Additionally, the model offers flexible control over the restoration process through textual prompts, vastly broadening the possibility of IR.
\cref{fig:teaser} illustrates the effects by our model.

Our work goes far beyond simply scaling.
While pursuing an increase in model scale, we face a series of complex challenges.
First, existing adaptor designs either too simple to meet the complex requirements of IR \cite{mou2023t2i} or are too large to train together with SDXL \cite{zhang2023adding}.
To solve this problem, we trim the ControlNet and designed a new connector called ZeroSFT to work with the pre-trained SDXL, aiming to efficiently implement the IR task while reducing computing costs.
In order to enhance the model's ability to accurately interpret the content of low-quality images, we fine-tune the image encoder to improve its robustness to variations in image degradation.
These measures make scaling the model feasible and effective, and greatly improve its stability.
Second, we collect 20 million high-quality, high-resolution images with descriptive text annotations, providing a solid foundation for the model's training.
We employ a counter-intuitive approach by integrating poor-quality samples into our training process. This allows us to enhance visual effects by utilizing prompts to guide the model away from negative qualities.
Finally, powerful generative prior is a double-edged sword.
Uncontrolled generation may reduce restoration fidelity, making IR no longer faithful to the input image.
To address the issue of low fidelity, we introduce the concept of restoration-guided sampling.
By integrating these strategies with efficient engineering practices, we not only facilitate the scaling up of SUPIR but also push the frontiers of advanced IR.

\vspace{-1mm}
\section{Related Work}
\label{sec:related work}

%%%%%%%%%%%%%%%%%%%%%%%%%%%%%%%%%%%%%%%%%%%%%%%%%%

\begin{figure*}[ht]
	%\newlength-4mm
	%\setlength{-4mm}{-0.4cm}
	\scriptsize
	\centering
    \includegraphics[width=\textwidth]{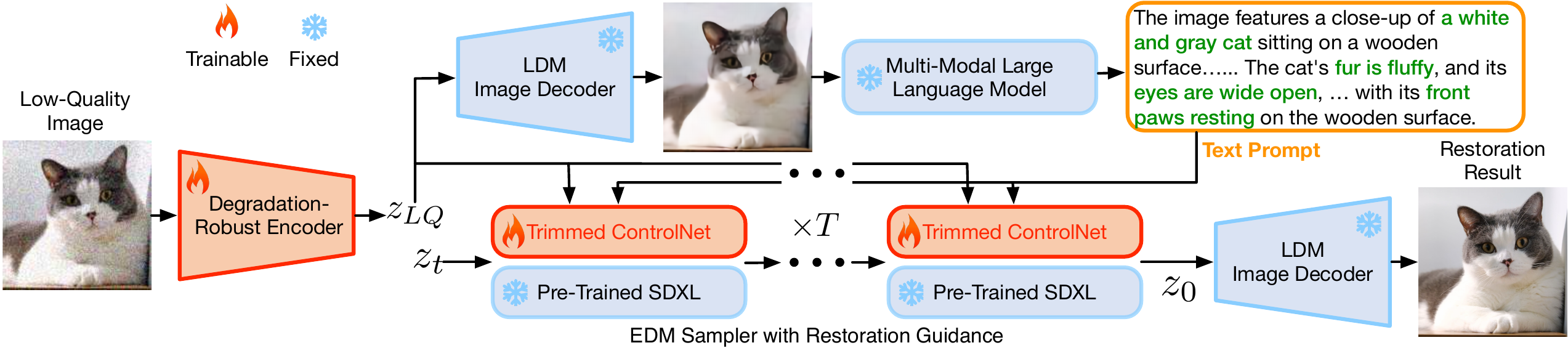}
	\vspace{-4mm}
	\caption{This figure briefly shows the workflow of the proposed SUPIR model.}
    \label{fig:pipeline}
	\vspace{-6mm}
\end{figure*}

%%%%%%%%%%%%%%%%%%%%%%%%%%%%%%%%%%%%%%%%%%%%%%%%%%
\vspace{-2mm}
\paragraph{Image Restoration.}
The goal of IR is to convert degraded images into high-quality degradation-free images \cite{zhang2017learning,gu2020pipal,zhang2019residual,zhang2020residual,zhang2022accurate,fan2020neural}.
In the early stage, researchers independently explored different types of image degradation, such as super-resolution (SR) \cite{dong2015image,dong2016accelerating,chen2023dual}, denoising \cite{zhang2017beyond,zhang2018ffdnet,chen2023masked}, and deblurring \cite{chen2023hierarchical,nah2017deep,tao2018scale}.
However, these methods are often based on specific degradation assumptions \cite{liu2022blind,michaeli2013nonparametric,gu2019blind} and therefore lack generalization ability to other degradations \cite{gu2023networks,zhang2023crafting,liu2021discovering,gu2024interpretability}.
Over time, the need for blind IR methods that are not based on specific degradation assumptions has grown \cite{gu2019blind,huang2020unfolding,liang2021flow,bell2019blind,hui2021learning,liang2022efficient,liang2021swinir,zhang2023practical,chen2022real}.
In this trend, some methods \cite{zhang2021designing,wang2021real} synthesize real-world degradation by more complex degradation models, and are well-known for handling multiple degradation with a single model.
DiffBIR \cite{lin2023diffbir} unifies different restoration problems into a single model.
In this paper, we adopt a similar setting to DiffBIR and use a single model to achieve effective processing of various severe degradations.

\vspace{-4mm}
\paragraph{Generative Prior.}
Generative priors are adept at capturing the inherent structures of the image, enabling the generation of images that follow natural image distribution.
The emergence of GANs \cite{goodfellow2014generative,radford2015unsupervised,karras2019style,karras2017progressive} has underscored the significance of generative priors in IR.
Various approaches employ generative priors, including GAN inversion \cite{gu2020image,menon2020pulse,abdal2019image2stylegan,pan2021exploiting,bau2020semantic}, GAN encoders \cite{zhu2022disentangled,chan2021glean}, or using GAN as the core module for IR \cite{yang2021gan,wang2021towards}.
Beyond GANs, other generative models can also serve as priors \cite{van2017neural,zhou2022towards,zhao2022rethinking,chen2022real,jo2021srflow,lugmayr2020learning,zhang2023unified}.
Our work primarily focuses on generative priors derived from diffusion models \cite{ho2020denoising,song2019generative,nichol2021glide,rombach2022high,ramesh2021zero,song2020score}, which excel in controllable generation \cite{mou2023t2i,zhang2023adding,choi2021ilvr,dhariwal2021diffusion,hu2021lora} and model scaling \cite{podell2023sdxl,ramesh2022hierarchical,saharia2022photorealistic}.
Diffusion models have also been effectively used as generative priors in IR \cite{kawar2022denoising,wang2022zero,lin2023diffbir,rombach2022high,wang2023exploiting}.
However, these diffusion-based IR methods' performance is constrained by the scale of the used generative models, posing challenges in further enhancing their effectiveness.

\vspace{-4mm}
\paragraph{Model Scaling} is an important means to further improve the capabilities of deep-learning models.
The most typical examples include the scaling of language models \cite{brown2020language,touvron2023llama,2023internlm}, text-to-image generation models \cite{saharia2022photorealistic,rombach2022high,podell2023sdxl,chen2023pixart,xue2023raphael,kang2023scaling}, and image segmentation models \cite{kirillov2023segment}.
The scale and complexity of these models have increased dramatically, now encompassing billions of parameters. This increase in parameters has also resulted in significant performance enhancements, showcasing the immense potential of model scaling \cite{kaplan2020scaling}.
However, scaling up is a systematic problem, involving model design, data collection, computing resources, and other limitations.
Many other tasks have not yet been able to enjoy the substantial performance improvements brought by scaling up. IR is one of them.

%%%%%%%%%%%%%%%%%%%%%%%%%%%%%%%%%%%%%%%%%%%%%%%%%%

\begin{figure*}[t]
    \centering
    \resizebox{0.76\textwidth}{!}{
        \includegraphics[width=\linewidth]{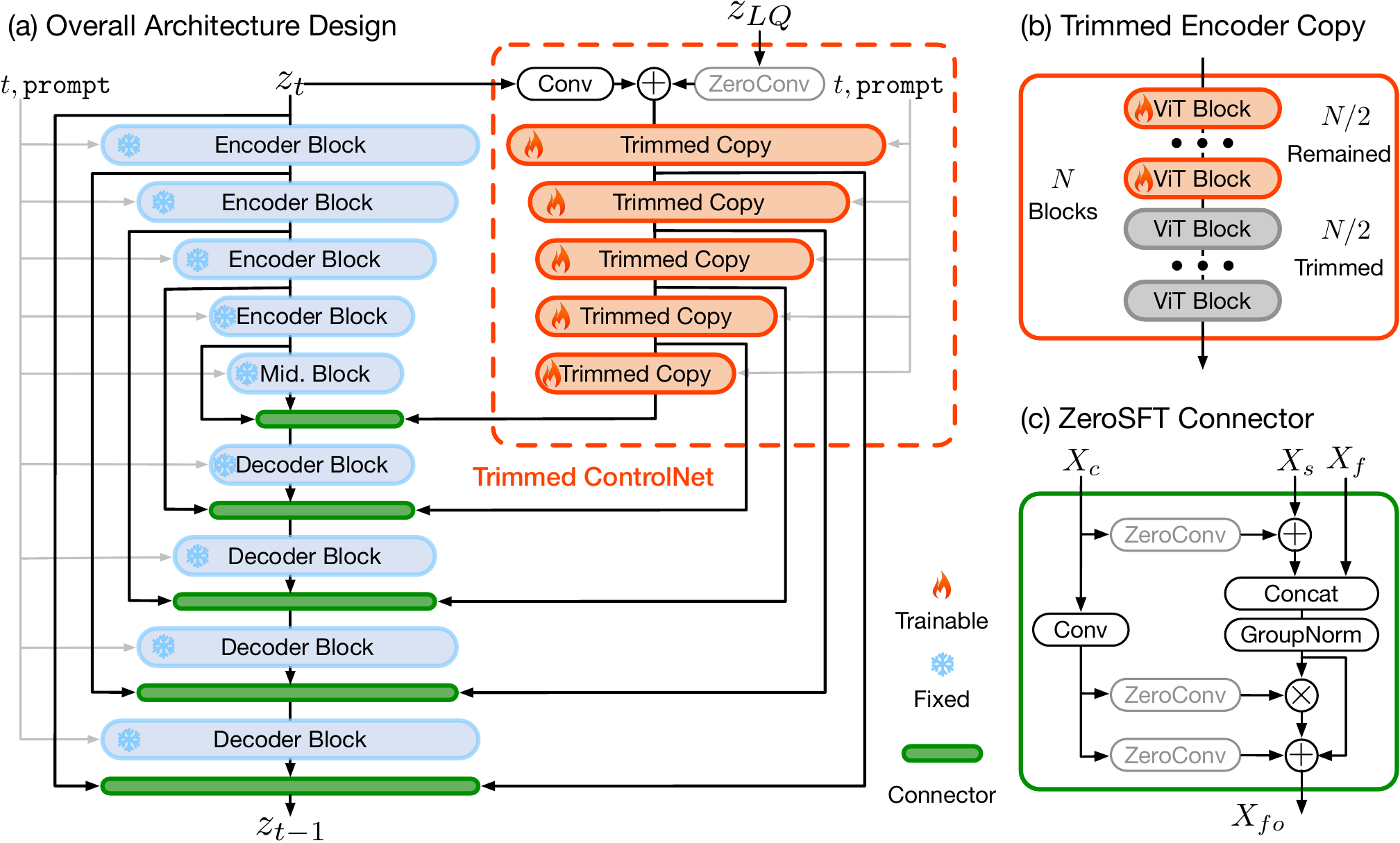}
    }
    \hfill 
    % \resizebox{0.35\textwidth}{!}{
    \raisebox{110pt}{
    \begin{minipage}[c]{0.20\textwidth}
        \caption{This figure illustrates (a) the overall architecture of the used SDXL and the proposed adaptor, (b) a trimmed trainable copy of the SDXL encoder with reduced ViT blocks for efficiency, and (c) a novel ZeroSFT connector for enhanced control in IR, where \(X_{f}\) and \(X_{s}\) denote the input feature maps from the Decoder and Encoder shortcut, respectively, \(X_{c}\) is the input from the adaptor, and \(X_{fo}\) is the output. The model is designed to effectively use the large-scale SDXL as a generative prior.}
	   \label{fig:arch}
    \end{minipage}
    }
    \vspace{-6mm}
\end{figure*}

%%%%%%%%%%%%%%%%%%%%%%%%%%%%%%%%%%%%%%%%%%%%%%%%%%
\vspace{-1mm}
\section{Method}
\label{sec:method}

\vspace{-2mm}
An overview of the proposed SUPIR method is shown in \cref{fig:pipeline}.
We introduce our method from three aspects:
\cref{sec:3.1} introduces our network designs and training method; \cref{sec:3.2} introduces the collection of training data and the introduction of textual modality; and \cref{sec:3.3} introduces the diffusion sampling method for IR.

\subsection{Model Scaling Up}
\label{sec:3.1}
\vspace{-1mm}
\paragraph{Generative Prior.}
There are not many choices for the large-scale generative models.
The only ones to consider are Imagen \cite{saharia2022photorealistic}, IF \cite{DeepFloyd2023}, and SDXL \cite{podell2023sdxl}.
Our selection settled on SDXL for the following reasons.
Imagen and IF prioritize text-to-image generation and rely on a hierarchical approach.
They first generate small-resolution images and then hierarchically upsample them.
SDXL aligns with our objectives by directly generating high-resolution images without a hierarchical design, effectively using its parameters to improve image quality rather than focusing on text interpretation.
Additionally, SDXL employs a \textit{Base}-\textit{Refine} strategy.
In the \textit{Base} model, diverse but lower-quality images are generated.
Subsequently, the \textit{Refine} model, utilizing training images of significantly higher quality but lesser diversity than those used by the \textit{Base} model, enhances the images' quality.
Given our approach of training with a vast dataset of high-quality images, the dual-phase design of SDXL becomes redundant for our objectives.
We opt for the \textit{Base} model, which has a greater number of parameters, making it an ideal generative prior.

\vspace{-4mm}
\paragraph{Degradation-Robust Encoder.}
In SDXL, the diffusion generation process is performed in the latent space.
The image is first mapped to the latent space through a pre-trained encoder.
To effectively utilize the pre-trained SDXL, our LQ image $x_{LQ}$ should also be mapped to the same latent space.
However, since the original encoder has not been trained on LQ images, using it for encoding will affect the model's judgment of LQ image content, and then misunderstand artifacts as image content \cite{lin2023diffbir}.
To this end, we fine-tune the encoder to make it robust to the degradation by minimizing:
$
    \mathcal{L}_{\mathcal{E}}=\|\mathcal{D}(\mathcal{E}_{\mathrm{dr}}(x_{LQ}))-\mathcal{D}(\mathcal{E}_{\mathrm{dr}}(x_{GT}))\|_2^2,
$
where $\mathcal{E}_{\mathrm{dr}}$ is the degradation-robust encoder to be fine-tuned, $\mathcal{D}$ is the fixed decoder, $x_{GT}$ is the ground truth.

\vspace{-4mm}
\paragraph{Large-Scale Adaptor Design.}
Considering the SDXL model as our chosen prior, we need an adaptor that can steer it to restore images according to the provided LQ inputs.
The Adaptor is required to identify the content in the LQ image and to finely control the generation at the pixel level.
LoRA \cite{hu2021lora}, T2I adaptor \cite{mou2023t2i}, and ControlNet \cite{zhang2023adding} are existing diffusion model adaptation methods, but none of them meet our requirements: LoRA limits generation but struggles with LQ image control; T2I lacks capacity for LQ image content identification; and ControlNet's direct copy is challenging for the SDXL model scale.
To address this issue, we design a new adaptor with two key features, as shown in \cref{fig:arch}(a).
First, we keep the high-level design of ControlNet but employ network trimming \cite{hu2016network} to directly trim some blocks within the trainable copy, achieving an engineering-feasible implementation.
Each block within the encoder module of SDXL is mainly composed of several Vision Transformer (ViT) \cite{dosovitskiy2020image} blocks.
We identified two key factors contributing to the effectiveness of ControlNet: large network capacity and efficient initialization of the trainable copy.
Notably, even partial trimming of blocks in the trainable copy retains these crucial characteristics in the adaptor.
Therefore, we simply trim half of the ViT blocks from each encoder block, as shown in \cref{fig:arch}(b).
Second, we redesign the connector that links the adaptor to SDXL.
While SDXL's generative capacity delivers excellent visual effects, it also renders pixel-level control challenging.
ControlNet employs zero convolution for generation guidance, but relying solely on residuals is insufficient for the control required by IR.
To amplify the influence of LQ guidance, we introduced a ZeroSFT module, as depicted in \cref{fig:arch}(c).
Building based on zero convolution, ZeroSFT encompasses an additional spatial feature transfer (SFT) \cite{wang2018recovering} operation and group normalization \cite{wu2018group}.

\vspace{-1mm}
\subsection{Scaling Up Training Data}
\label{sec:3.2}
\vspace{-1mm}
\paragraph{Image Collection.}
The scaling of the model requires a corresponding scaling of the training data \cite{kaplan2020scaling}.
But there is no large-scale high-quality image dataset available for IR yet.
Although DIV2K \cite{Agustsson_2017_CVPR_Workshops} and LSDIR \cite{ETHZ_Yawli} offer high image quality, they are limited in quantity.
Larger datasets like ImageNet (IN) \cite{deng2009imagenet}, LAION-5B \cite{schuhmann2022laion}, and SA-1B \cite{kirillov2023segment} contain more images, but their image quality does not meet our high standards. 
To this end, we collect a large-scale dataset of high-resolution images, which includes 20 million 1024$\times$1024 high-quality, texture-rich images.
A comparison on the scales of the collected dataset and the existing dataset is shown in \cref{fig:arch}.
We also included an additional 70K unaligned high-resolution facial images from the FFHQ-raw dataset \cite{karras2019style} to improve the model's face restoration performance.
In \cref{fig:alg}(a), we show the relative size of our data compared to other well-known datasets.

\begin{figure}[t]
    \centering
    \includegraphics[width=\linewidth]{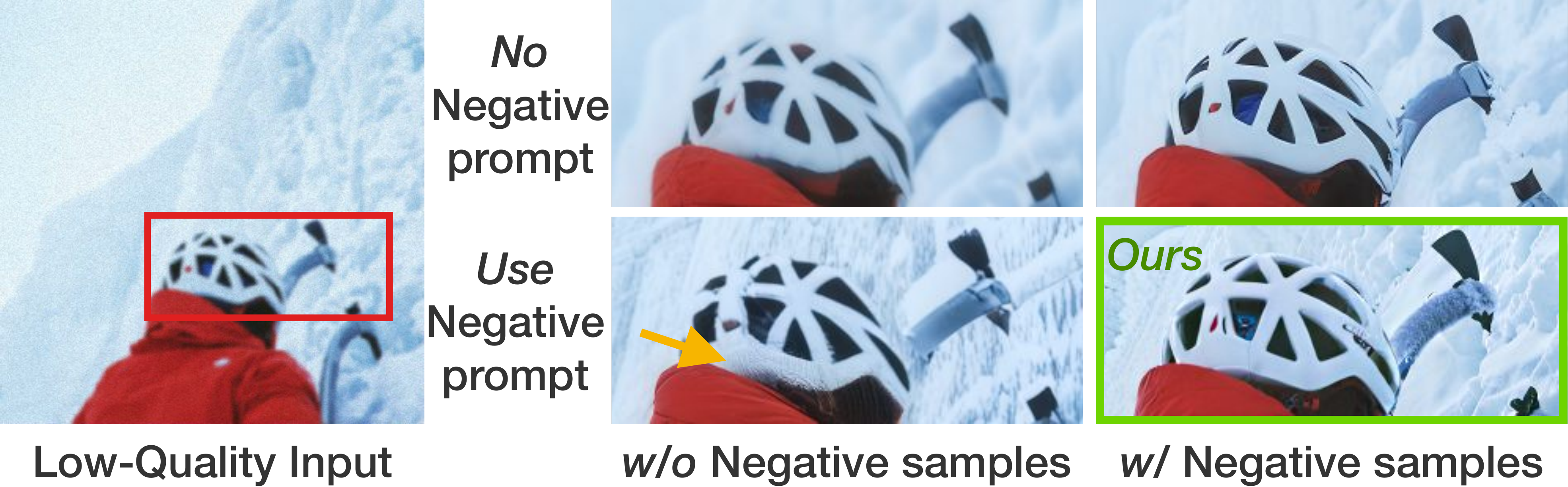}
    \vspace{-6mm}
    \caption{
    CFG introduces artifacts without negative training samples, hindering visual quality improvement. Adding negative samples allows further quality enhancement through CFG.
    }
    \label{fig:neg}
    \vspace{-6mm}
\end{figure}
\vspace{-4mm}
\paragraph{Multi-Modality Language Guidance.}
Diffusion models are renowned for their ability to generate images based on textual prompts.
We believe that textual prompts can also aid IR:
(1) Understanding image content is crucial for IR.
Existing frameworks often overlook or implicitly handle this understanding \cite{gu2021interpreting,gu2023networks}.
By incorporating textual prompts, we explicitly convey the understanding of LQ images to the IR model, facilitating targeted restoration of missing information.
(2) In cases of severe degradation, even the best IR models struggle to recover completely lost information.
In such cases, textual prompts can serve as a control mechanism, enabling targeted completion of missing information based on user preferences.
(3) We can also describe the desired image quality through text, further enhancing the perceptual quality of the output.
See \cref{fig:teaser}(b) for some examples.
To this end, we make two main modifications.
First, we revise the overall framework to incorporate the LLaVA multi-modal LLM \cite{liu2023llava} into our pipeline, as shown in \cref{fig:pipeline}.
LLaVA takes the degradation-robust processed LQ images $x_{LQ}'=\mathcal{D}(\mathcal{E}_{\mathrm{dr}}(x_{LQ}))$ as input and explicitly understands the content within the images, outputting in the form of textual descriptions.
These descriptions are then used as prompts to guide the restoration.
This process can be automated during testing, eliminating the need for manual intervention.
Secondly, following the approach of PixART \cite{chen2023pixart}, we also collect textual annotations for all the training images, to reinforce the role of textual control during the training of out model.
These two changes endow SUPIR with the ability to understand image content and to restore images based on textual prompts.

\vspace{-4mm}
\paragraph{Negative-Quality Samples and Prompt.}
Classifier-free guidance (CFG) \cite{ho2022classifier} provides another way of control by using negative prompts to specify undesired content for the model.
We can use this feature to specify the model NOT to produce low-quality images.
Specifically, at each step of diffusion, we will make two predictions using positive prompts $\mathrm{pos}$ and negative prompts $\mathrm{neg}$, and take the fusion of these two results as the final output $z_{t-1}$:
\begin{align}
    &z_{t-1}^{\mathrm{pos}}=\mathcal{H}(z_t,z_{LQ},\sigma_t,\mathrm{pos}),
    z_{t-1}^{\mathrm{neg}}=\mathcal{H}(z_t,z_{LQ},\sigma_t,\mathrm{neg}),\notag\\
    &z_{t-1}=z_{t-1}^{\mathrm{pos}}+\lambda_{\mathrm{cfg}}\times(z_{t-1}^{\mathrm{pos}}-z_{t-1}^{\mathrm{neg}}),\notag
\end{align}
where $\mathcal{H}(\cdot)$ is our diffusion model with adaptor, $\sigma_t$ is the variance of the noise at time-step $t$, and $\lambda_{\mathrm{cfg}}$ is a hyper-parameter.
In our framework, $\mathrm{pos}$ can be the image description with positive words of quality, and $\mathrm{neg}$ is the negative words of quality, \eg, ``\textit{oil painting, cartoon, blur, dirty, messy, low quality, deformation, low resolution, over-smooth}''.
Accuracy in predicting both positive and negative directions is crucial for the CFG technique.
However, the absence of negative-quality samples and prompts in our training data may lead to a failure of the fine-tuned SUPIR in understanding negative prompts.
Therefore, using negative-quality prompts during sampling may introduce artifacts, see \cref{fig:neg} for an example.
To address this problem, we used SDXL to generate 100K images corresponding to the negative-quality prompts.
We counter-intuitively add these low-quality images to the training data to ensure that negative-quality concept can be learned by the proposed SUPIR model.

\vspace{-1mm}
\subsection{Restoration-Guided Sampling}
\label{sec:3.3}
\begin{figure}[t]
    \vspace{-3mm}
    \centering
    \includegraphics[width=0.59\linewidth]{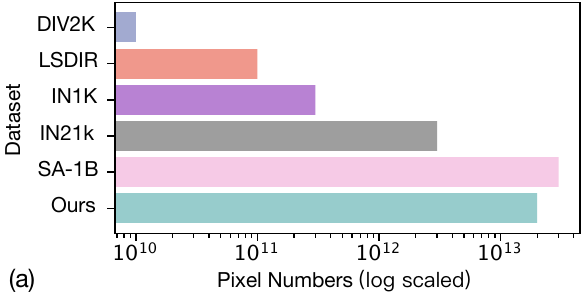}
    \includegraphics[width=0.39\linewidth]{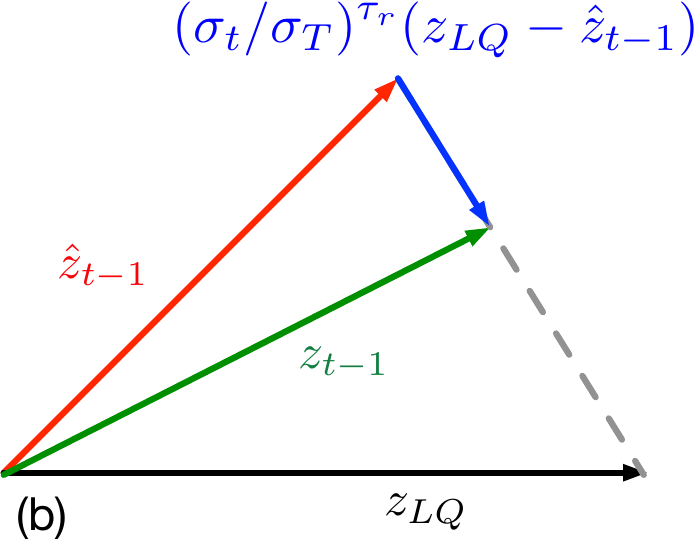}
    \vspace{-3mm}
    \caption{
    (a) We show the relative size of our data compared to other well-known datasets. Compared with SA-1B \cite{kirillov2023segment}, our dataset has higher quality and more image diversity. (b) We demonstrate our restoration-guided sampling mechanism.
    }
    \label{fig:alg}
    \vspace{-6mm}
\end{figure}

\vspace{-2mm}
Powerful generative prior is a double-edged sword, as too much generation capacity will in turn affect the fidelity of the recovered image.
This highlights the fundamental difference between IR tasks and generation tasks.
We need means to limit the generation to ensure that the image recovery is faithful to the LQ image.
We modified the EDM sampling method \cite{karras2022elucidating} and proposed a restoration-guided sampling method to solve this problem.
We hope to selectively guide the prediction results $z_{t-1}$ to be close to the LQ image $z_{LQ}$ in each diffusion step.
The specific algorithm is shown in \cref{alg:stochastic}, where $T$ is the total step number, $\{\sigma_t\}_{t=1}^T$ are the noise variance for $T$ steps, $c$ is the additional text prompt condition.
$\tau_r$, $S_{\mathrm{churn}}$, $S_{\mathrm{noise}}$, $S_{\mathrm{min}}$, $S_{\mathrm{max}}$ are five hyper-parameters, but only $\tau_r$ is related to the restoration guidance, the others remain unchanged compared to the original EDM method \cite{karras2022elucidating}.
For better understanding, a simple diagram is shown in \cref{fig:alg}(b).
We perform weighted interpolation between the predicted output $\hat{z}_{t-1}$ and the LQ latent $z_{LQ}$ as the restoration-guided output $z_{t-1}$.
Since the low-frequency information of the image is mainly generated in the early stage of diffusion prediction \cite{rombach2022high} (where $t$ and $\sigma_t$ are relatively large, and the weight $k=(\sigma_t/\sigma_T)^{\tau_r}$ is also large), the prediction result is closer to $z_{LQ}$ to enhance fidelity.
In the later stages of diffusion prediction, mainly high-frequency details are generated.
There should not be too many constraints at this time to ensure that detail and texture can be adequately generated.
At this time, $t$ and $\sigma_t$ are relatively small, and weight $k$ is also small.
Therefore, the predicted results will not be greatly affected
Through this method, we can control the generation during the diffusion sampling process to ensure fidelity.

\begin{algorithm}[t]
\footnotesize
\captionof{algorithm}[stochastic]{Restoration-Guided Sampling.}
\hspace*{\algorithmicindent} \textbf{Input:} $\mathcal{H}$, $\{\sigma_t\}_{t=1}^T$, $z_{LQ}$, $c$ \\
\hspace*{\algorithmicindent} \textbf{Hyper-parameter:} $\tau_r$, $S_{\mathrm{churn}}$, $S_{\mathrm{noise}}$, $S_{\mathrm{min}}$, $S_{\mathrm{max}}$
% \hspace*{\algorithmicindent} \textbf{Output:} $z_0$
\begin{algorithmic}[1]
  % \Procedure{Sampler}{}
  \State{{\bf sample} $z_T \sim \mathcal{N}(\mathbf{0}, \sigma_T^2 \mathbf{I})$}
  \For{$t \in \{T, \dots, 1\}$}
  \State{{\bf sample} $\boldsymbol{\epsilon}_t \sim \mathcal{N}\left(\mathbf{0}, S_{\mathrm{noise}}^2 \mathbf{I}\right)$}
  \State{$\gamma_t\leftarrow \begin{cases}\min \left(\frac{S_{\mathrm{churn}}}{N}, \sqrt{2}-1\right) & \text { if } \sigma_t \in\left[S_{\mathrm{min}}, S_{\mathrm{max}}\right] \\ 0 & \text { otherwise }\end{cases}$}
  % \State{$\hat{\sigma}_t \leftarrow \sigma_t+\gamma_t \sigma_t$}
  \State{$k_t \leftarrow (\sigma_t/\sigma_T)^{\tau_r}$, $\hat{z}_t \leftarrow z_t+\sqrt{\hat{\sigma}_t^2-\sigma_t^2} \boldsymbol{\epsilon}_t$, $\hat{\sigma}_t \leftarrow \sigma_t+\gamma_t \sigma_t$}
  % \State{$k_t \leftarrow (\sigma_t/\sigma_T)^{\tau_r}$}
  \State{$\hat{z}_{t-1}\leftarrow \mathcal{H}\left(\hat{z}_t,z_{LQ},\hat{\sigma}_t,c\right)$}
  \State{$d_t\leftarrow(\hat{z}_t-(
  % \textcolor{red}{(1-k_t)\mathcal{H}\left(\hat{z}_t,z_{LQ},\hat{\sigma}_t,c\right)} + \textcolor{red}{k_t z_{LQ}})
  \textcolor{red}{\hat{z}_{t-1}}+k_t(\textcolor{red}{z_{LQ}}-\textcolor{red}{\hat{z}_{t-1}}))
  )/ \hat{\sigma}_t$}
  \State{$z_{t-1} \leftarrow \hat{z}_t+\left(\sigma_{t-1}-\hat{\sigma}_t\right) d_t$}
  \EndFor
  % \EndProcedure
\end{algorithmic}
\label{alg:stochastic}
\end{algorithm}

\begin{figure*}[]
	%\newlength-4mm
	%\setlength{-4mm}{-0.4cm}
	\scriptsize
	\centering
    \vspace{-4mm}
    \resizebox{\textwidth}{!}{
	\begin{tabular}{l}
		\hspace{-0.42cm}
		\begin{adjustbox}{valign=t}
			\begin{tabular}{c}
				\includegraphics[width=0.409\textwidth]{{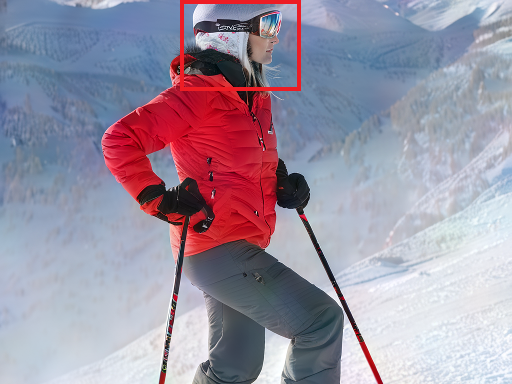}}
                \\
                \method{} (ours)
				\hspace{-10mm}
			\end{tabular}
		\end{adjustbox}
		\hspace{-5mm}
		\begin{adjustbox}{valign=t}
			\begin{tabular}{ccc}
				\includegraphics[width=0.19\textwidth]{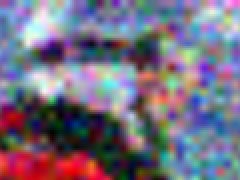} \hspace{-4.5mm} &
				\includegraphics[width=0.19\textwidth]{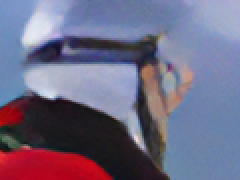} \hspace{-4.5mm} &
				\includegraphics[width=0.19\textwidth]{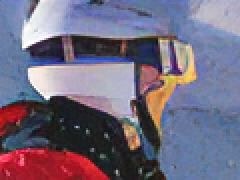} \hspace{-4.5mm} 
				\\
				Low-Quality Input \hspace{-2mm} &
				Real-ESRGAN+\cite{wang2021real} \hspace{-3.5mm} &
				StableSR\cite{wang2023exploiting} \hspace{-3.5mm}
				\\
				\includegraphics[width=0.19\textwidth]{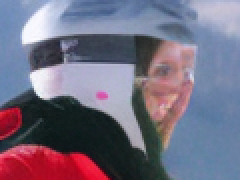} \hspace{-4.5mm} &
				\includegraphics[width=0.19\textwidth]{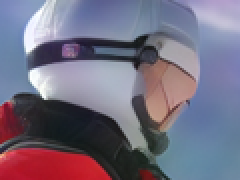} \hspace{-4.5mm} &
				\includegraphics[width=0.19\textwidth]{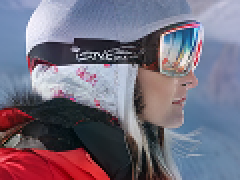} \hspace{-4.5mm}
				\\ 
				DiffBIR\cite{lin2023diffbir} \hspace{-3.5mm} &
				PASD\cite{yang2023pixel} \hspace{-3.5mm} &
				\method{} (ours) \hspace{-3.5mm}
			\end{tabular}
		\end{adjustbox}
		\hspace{-2mm}
        \\
		\hspace{-0.42cm}
		\begin{adjustbox}{valign=t}
			\begin{tabular}{c}
				\includegraphics[width=0.409\textwidth]{{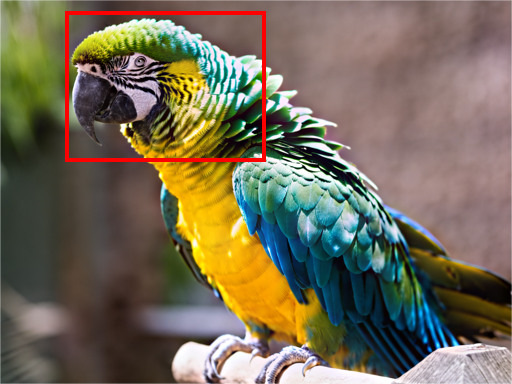}}
                \\
                \method{} (ours).
				\hspace{-10mm}
			\end{tabular}
		\end{adjustbox}
		\hspace{-5mm}
		\begin{adjustbox}{valign=t}
			\begin{tabular}{ccc}
				\includegraphics[width=0.19\textwidth]{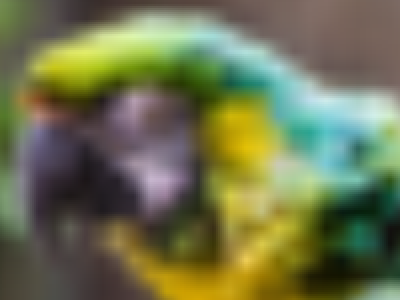} \hspace{-4.5mm} &
				\includegraphics[width=0.19\textwidth]{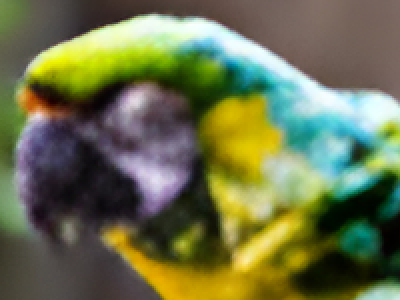} \hspace{-4.5mm} &
				\includegraphics[width=0.19\textwidth]{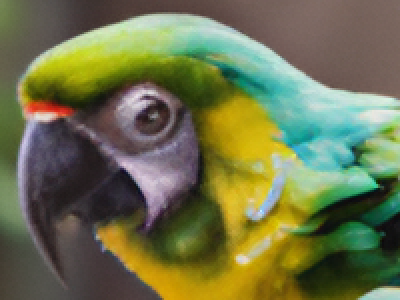} \hspace{-4.5mm} 
				\\
				Low-Quality Input \hspace{-2mm} &
				Real-ESRGAN+\cite{wang2021real} \hspace{-3.5mm} &
				StableSR\cite{wang2023exploiting} \hspace{-3.5mm}
				\\
				\includegraphics[width=0.19\textwidth]{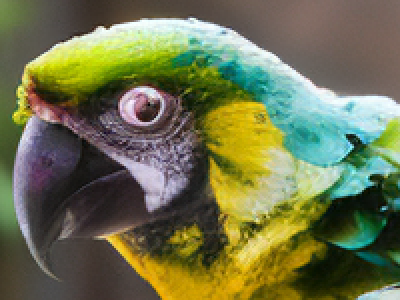} \hspace{-4.5mm} &
				\includegraphics[width=0.19\textwidth]{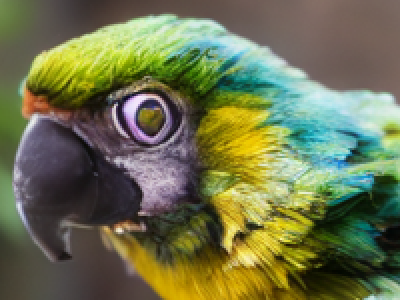} \hspace{-4.5mm} &
				\includegraphics[width=0.19\textwidth]{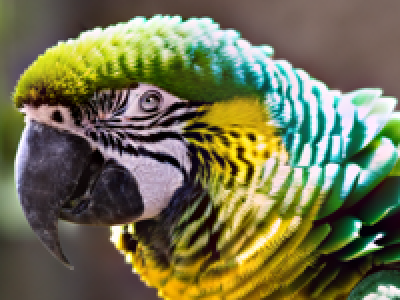} \hspace{-4.5mm}
				\\ 
				DiffBIR\cite{lin2023diffbir} \hspace{-3.5mm} &
				PASD\cite{yang2023pixel} \hspace{-3.5mm} &
				\method{} (ours) \hspace{-3.5mm}
			\end{tabular}
		\end{adjustbox}
		\hspace{-2mm}
	\end{tabular}
    }
    % \hspace{-8mm}
    \vspace{-4mm}
	\caption{Qualitative comparison with different methods. Our method can accurately restore the texture and details of the corresponding object under challenging degradation. Other methods fail to recover semantically correct details such as broken beaks and irregular faces.}
    \label{fig1:main_visual}
	\vspace{-4mm}
\end{figure*}
% %%%%%%%%%%%%%%%%%%%%%%%%%%%%%%%%%%%%%%%%%%%%%%%%%%
\begin{figure}[t]
    \centering
    \includegraphics[width=\linewidth]{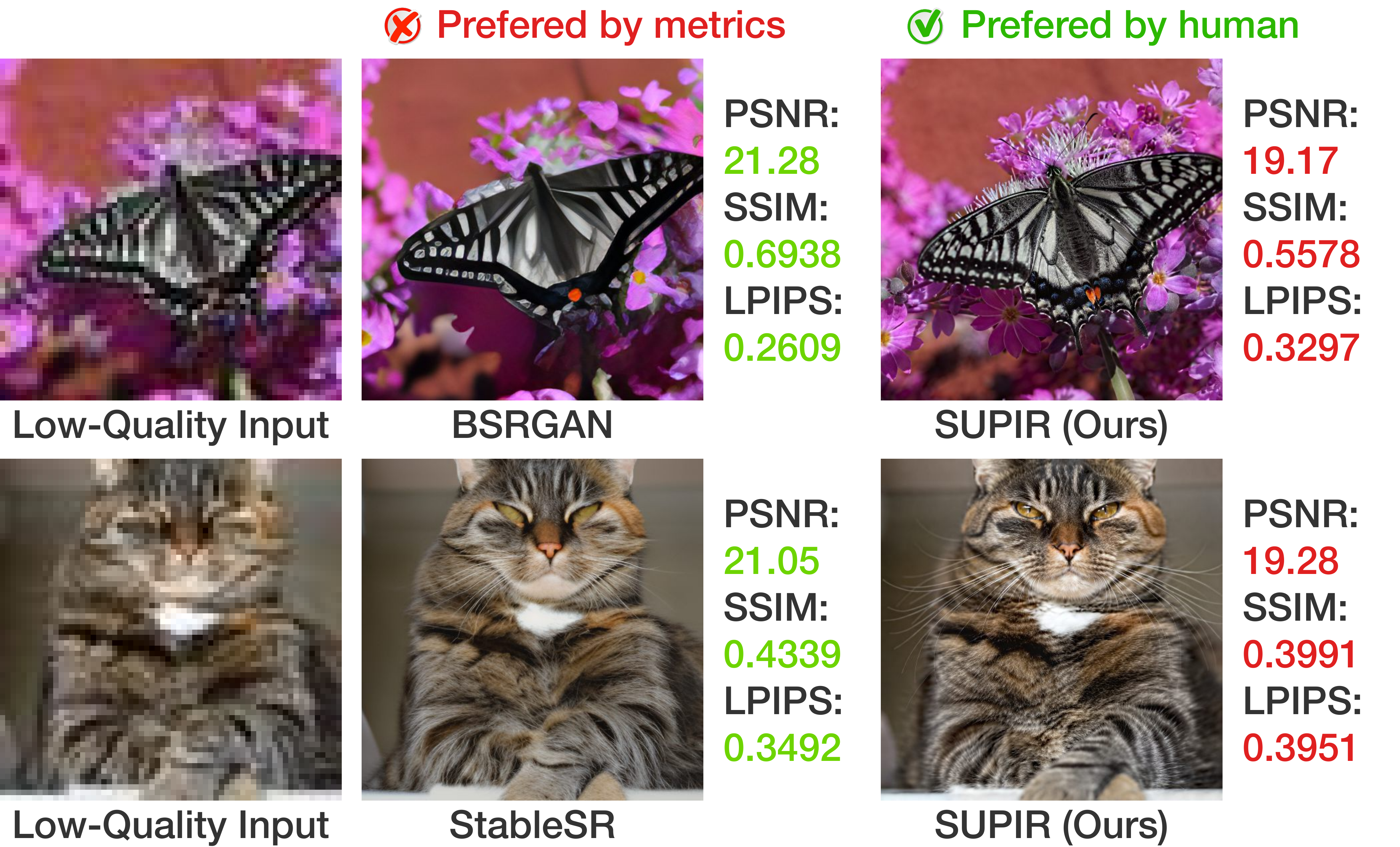}
    \vspace{-7mm}
    \caption{
    These examples show the misalignment between metric evaluation and human evaluation. \method{} generates images with high-fidelity textures, but obtains lower metrics.
    }
    \label{fig:metric}
    \vspace{-6mm}
\end{figure}
\vspace{-1mm}
\section{Experiments}
\label{sec:experiments}
\vspace{-1mm}
\subsection{Model Training and Sampling Settings}
\vspace{-2mm}
For training, the overall training data includes 20 million high-quality images with text descriptions, 70K face images and 100K negative-quality samples, together their corresponding prompts.
To enable a larger batch size, we crop images into 512$\times$512 patches during training.
We train our model using a synthetic degradation model, following the setting used by Real-ESRGAN \cite{wang2021real}, the only difference is that we resize the produced LQ images to 512$\times$512 for training.
We use the AdamW optimizer \cite{loshchilov2017decoupled} with a learning rate of $0.00001$.
The training process spans 10 days and is conducted on 64 Nvidia A6000 GPUs, with a batch size of 256.
For testing, the hyper-parameters are $T$=100, $\lambda_{\mathrm{cfg}}$=7.5, and $\tau_r=4$.
Our method is able to process images with the size of 1024$\times$1024.
We resize the short side of the input image to 1024 and crop a 1024$\times$1024 sub-image for testing, and then resize it back to the original size after restoration.
Unless stated otherwise, prompts will not be provided manually -- the processing will be entirely automatic.

\vspace{-1mm}
\subsection{Comparison with Existing Methods}
\label{sec:existing_methods}
\vspace{-2mm}
Our method can handle a wide range of degradations, and we compare it with the latest methods with the same capabilities, including BSRGAN \cite{zhang2021designing}, Real-ESRGAN \cite{wang2021real}, StableSR \cite{wang2023exploiting}, DiffBIR \cite{lin2023diffbir} and PASD \cite{yang2023pixel}.
Some of them are constrained to generating images of 512$\times$512 size.
In our comparison, we crop the test image to meet this requirement and downsample our results.
We conduct comparisons on both synthetic data and real-world data.

\begin{table}[]
    \centering
    \hspace{-2.5mm}
    \scalebox{0.6}{
    \begin{tabular}{p{1.9cm}p{2.2cm}cccccc}
    \toprule
        Degradation & Method & PSNR & SSIM & LPIPS$\downarrow$ & ManIQA & ClipIQA & MUSIQ \\
        \midrule
        \multirow{6}{*}{\begin{minipage}{2cm}Single:\\SR ($\times4$)\end{minipage}} 
        & BSRGAN & \rf{25.06} & \rf{0.6741} & 0.2159 & 0.2214 & 0.6169 & 70.38\\
        & Real-ESRGAN & 24.26 & \bd{0.6657} & \bd{0.2116} & 0.2287 & 0.5884 & 69.51\\
        & StableSR & 22.59 & 0.6019 & 0.2130 & \bd{0.3304} & \bd{0.7520} & \bd{72.94}\\
        & DiffBIR & 23.44 & 0.5841 & 0.2337 & 0.2879 & 0.7147 & 71.64\\
        & PASD & \bd{24.90} & 0.6653 & \rf{0.1893} & 0.2607 & 0.6466 & 71.39\\
        & SUPIR (ours) & 22.66 & 0.5763 & 0.2662 & \rf{0.4738} & \rf{0.8049} & \rf{73.83}\\
        \midrule
        \multirow{6}{*}{\begin{minipage}{2cm}Single:\\SR ($\times8$)\end{minipage}} 
        & BSRGAN & \rf{22.26} & \bd{0.5212} & 0.3523 & 0.2069 & 0.5836 & 67.04\\
        & Real-ESRGAN & 21.79 & \rf{0.5280} & 0.3276 & 0.2051 & 0.5349 & 63.80\\
        & StableSR & 21.27 & 0.4857 & 0.3118 & \bd{0.3039} & \bd{0.7333} & \bd{71.74}\\
        & DiffBIR & 21.86 & 0.4957 & \bd{0.3106} & 0.2845 & 0.7080 & 70.26\\
        & PASD & \bd{21.97} & 0.5149 & \rf{0.3034} & 0.2412 & 0.6402 & 70.20\\
        & SUPIR (ours)  & 20.68 & 0.4488 & 0.3749 & \rf{0.4687} & \rf{0.8009} & \rf{73.16}\\
        \midrule
        \multirow{6}{*}{\begin{minipage}{2cm}Mixture:\\Blur ($\sigma$=2) $+$\\SR ($\times4$)\end{minipage}} 
        & BSRGAN & \rf{24.97} & \rf{0.6572} & 0.2261 & 0.2127 & 0.5984 & 69.44\\
        & Real-ESRGAN & 24.08 & 0.6496 & \bd{0.2208} & 0.2357 & 0.5853 & 69.27\\
        & StableSR & 22.26 & 0.5721 & 0.2301 & \bd{0.3204} & \bd{0.7488} & \bd{72.87}\\
        & DiffBIR & 23.28 & 0.5741 & 0.2395 & 0.2829 & 0.7055 & 71.22\\
        & PASD & \bd{24.85} & \bd{0.6560} & \rf{0.1952} & 0.2500 & 0.6335 & 71.07\\
        & SUPIR (ours)  & 22.43 & 0.5626 & 0.2771 & \rf{0.4757} & \rf{0.8110} & \rf{73.55}\\
        \midrule
        \multirow{6}{*}{\begin{minipage}{2cm}Mixture:\\SR ($\times4$)$+$\\Noise ($\sigma$=40)\end{minipage}} 
        & BSRGAN & 17.74 & 0.3816 & 0.5659 & 0.1006 & 0.4166 & 51.25\\
        & Real-ESRGAN & 21.46 & \rf{0.5220} & 0.4636 & 0.1236 & 0.4536 & 52.23\\
        & StableSR & 20.88 & 0.4174 & 0.4668 & 0.2365 & 0.5833 & 63.54\\
        & DiffBIR & \rf{22.08} & 0.4918 & \rf{0.3738} & 0.2403 & \bd{0.6435} & 65.97\\
        & PASD & \bd{21.79} & \bd{0.4983} & \bd{0.3842} & \bd{0.2590} & 0.5939 & \bd{69.09}\\
        & SUPIR (ours)  & 20.77 & 0.4571 & 0.3945 & \rf{0.4674} & \rf{0.7840} & \rf{73.35}\\
        \midrule
        \multirow{6}{*}{\begin{minipage}{2cm}Mixture:\\Blur ($\sigma$=2) $+$\\SR ($\times4$)$+$\\Noise ($\sigma$=20)$+$\\JPEG ($q$=50)\end{minipage}} 
        & BSRGAN & \rf{22.88} & \rf{0.5397} & \bd{0.3445} & 0.1838 & 0.5402 & 64.81\\
        & Real-ESRGAN & \bd{22.01} & \bd{0.5332} & 0.3494 & 0.2115 & 0.5730 & 64.76\\
        & StableSR & 21.39 & 0.4744 & \rf{0.3422} & \bd{0.2974} & \bd{0.7354} & \bd{70.94}\\
        & DiffBIR & 21.79 & 0.4895 & 0.3465 & 0.2821 & 0.7059 & 69.28\\
        & PASD & 21.90 & 0.5118 & 0.3493 & 0.2397 & 0.6326 & 70.43\\
        & SUPIR (ours)  & 20.84 & 0.4604 & 0.3806 & \rf{0.4688} & \rf{0.8021} & \rf{73.58}\\
        \bottomrule
    \end{tabular}}
    \vspace{-3mm}
    \caption{Quantitative comparison. \rf{Red} and \bd{blue} colors represent the best and second best performance. $\downarrow$ represents the smaller the better, and for the others, the bigger the better.}
    \label{tab:synthetic_comp}
    \vspace{-2mm}
\end{table}

\begin{table}
    \centering
    \hspace{-4mm}
    \subfloat[Quantitative comparison on 60 real-world LQ images.\label{tab:real}]{
    \scalebox{0.6}{
    \begin{tabular}{cccccccc}
                \toprule
                Metrics & BSRGAN & Real-ESRGAN & StableSR&DiffBIR& PASD& Ours\\
                \midrule
                CLIP-IQA & 0.4119 & 0.5174 & 0.7654 & 0.6983 & \bd{0.7714} & \rf{0.8232} \\
                MUSIQ  & 55.64 & 59.42 & 70.70  & 69.69  & \bd{71.87}  & \rf{73.00}  \\
                MANIQA   & 0.1585 & 0.2262 & 0.3035 & 0.2619 & \bd{0.3169} & \rf{0.4295} \\
                \bottomrule
  \end{tabular}
    }
    }
    \\
    \hspace{-4.5mm}
    \subfloat[Ablation study of quality prompts and negative training samples.\label{tab:prompts}]{
    \scalebox{0.6}{
    \begin{tabular}{ccccccccc}
    \toprule
        Negative & \multicolumn{2}{c}{Prompts} & \multirow{2}{*}{PSNR} & \multirow{2}{*}{SSIM} & \multirow{2}{*}{LPIPS$\downarrow$} & \multirow{2}{*}{ManIQA} & \multirow{2}{*}{ClipIQA} & \multirow{2}{*}{MUSIQ} \\
        Samples & Positive & Negative & \\
        \midrule
         \checkmark & & & 22.90  & 0.5519 & 0.3010 & 0.3129 & 0.7049   & 68.94 \\
         \checkmark & \checkmark &  & 22.31  & 0.5250 & 0.3108 & 0.4018 & 0.7937   & 72.00\\
         \checkmark & & \checkmark & 20.63  & 0.4747 & 0.3603 & 0.4678 & 0.7933   & 73.60\\
         \checkmark & \checkmark & \checkmark & 20.66  & 0.4763 & 0.3412 & 0.4740 & 0.8164   & 73.66\\
         & \checkmark & \checkmark & 21.79 & 0.5119 & 0.3139 & 0.3180 & 0.7102 & 72.68 \\
    \bottomrule
    \end{tabular}
    }
    }
    \\
    \hspace{-4.5mm}
    \subfloat[Ablation study of zero convolution and the proposed ZeroSFT.\label{tab:connectors}]{
    \scalebox{0.6}{
    \begin{tabular}{p{4.45cm}cccccc}
    \toprule
        Connector & PSNR & SSIM & LPIPS$\downarrow$ & ManIQA & ClipIQA & MUSIQ \\
        \midrule
        Zero Convolution \cite{zhang2023adding} & 19.47 & 0.4261 & 0.3969 & 0.4845 & 0.8184   & 74.00 \\
        ZeroSFT & 20.66 & 0.4763 & 0.3412 & 0.4740 & 0.8164   & 73.66\\
         \bottomrule
    \end{tabular}
    }
    }
    % \\
    % \hspace{-4.5mm}
    % \subfloat[Ablation study of different training data.\label{tab:data}]{
    % \scalebox{0.6}{
    % \begin{tabular}{p{2.1cm}p{2cm}cccccc}
    % \toprule
    %     Dataset & Scale & PSNR & SSIM & LPIPS$\downarrow$ & MANIQA & CLIP-IQA & MUSIQ \\
    %     \midrule
    %     DIV2K \cite{Agustsson_2017_CVPR_Workshops} & & 21.62 & 0.5000 & 0.4750 & 0.1001 & 0.4288 & 52.39 \\
    %     LSDIR \cite{ETHZ_Yawli} & & 21.40 & 0.4800 & 0.4808 & 0.1830 & 0.5384 & 57.32\\
    %     The collected & & 21.09 & 0.4933 & 0.3523 & 0.3905 & 0.7646 & 73.50  \\
    %     \bottomrule
    % \end{tabular}
    % }
    % }
    \vspace{-3mm}
    \caption{Real-world comparison results and ablation studies.}
    \vspace{-4mm}
\end{table}

% \begin{table}[]
%     \centering
%     \begin{tabular}{c|cccccc}
%     \hline
%     $\tau_r$ & PSNR   & SSIM   & LPIPS  & MANIQA & CLIP-IQA & MUSIQ\\
%     \hline
%     1        & 22.03 & 0.5273 & 0.3281 & 0.4108 & 0.7647  & 70.57 \\
%     2        & 21.02 & 0.4894 & 0.3339 & 0.4690 & 0.8151  & 73.56 \\
%     3        & 20.72 & 0.4804 & 0.3361 & 0.4742 & 0.8112  & 73.58 \\
%     4        & 20.66 & 0.4763 & 0.3412 & 0.4740 & 0.8164  & 73.66 \\
%     5        & 20.59 & 0.4728 & 0.3446 & 0.4802 & 0.8174  & 73.71 \\
%     6        & 20.58 & 0.4739 & 0.3387 & 0.4770 & 0.8227  & 73.69 \\
%     N/A      & 20.40 & 0.4675 & 0.3468 & 0.4760 & 0.8179  & 73.62 \\
%     \hline
%     \end{tabular}
%     \caption{Caption}
%     \label{tab:my_label}
% \end{table}

\vspace{-5mm}
\paragraph{Synthetic Data.}
To synthesize LQ images for testing, we follow previous works \cite{kong2022reflash,zhang2023crafting} and demonstrate our effects on several representative degradations, including both single degradations and complex mixture degradations.
Specific details can be found in \cref{tab:synthetic_comp}.
We selected the following metrics for quantitative comparison: full-reference metrics PSNR, SSIM, LPIPS \cite{zhang2018unreasonable}, and the non-reference metrics ManIQA \cite{yang2022maniqa}, ClipIQA \cite{wang2023exploring}, MUSIQ \cite{ke2021musiq}.
It can be seen that our method achieves the best results on all non-reference metrics, which reflects the excellent image quality of our results.
At the same time, we also note the disadvantages of our method in full-reference metrics.
We present a simple experiment that highlights the limitations of these full-reference metrics, see \cref{fig:metric}.
It can be seen that our results have better visual effects, but they do not have an advantage in these metrics.
This phenomenon has also been noted in many studies as well \cite{blau2018perception,gu2020pipal,gu2022ntire}.
We argue that with the improving quality of IR, there is a need to reconsider the reference values of existing metrics and suggest more effective ways to evaluate advanced IR methods.
We also show some qualitative comparison results in \cref{fig1:main_visual}.
Even under severe degradation, our method consistently produces highly reasonable and high-quality images that faithfully represent the content of the LQ images.

%%%%%%%%%%%%%%%%%%%%%%%%%%%%%%%%%%%%%%%%%%%%%%%%%%

\begin{figure}[t]
	%\newlength-4mm
	%\setlength{-4mm}{-0.4cm}
	\scriptsize
	\centering
    \includegraphics[width=\linewidth]{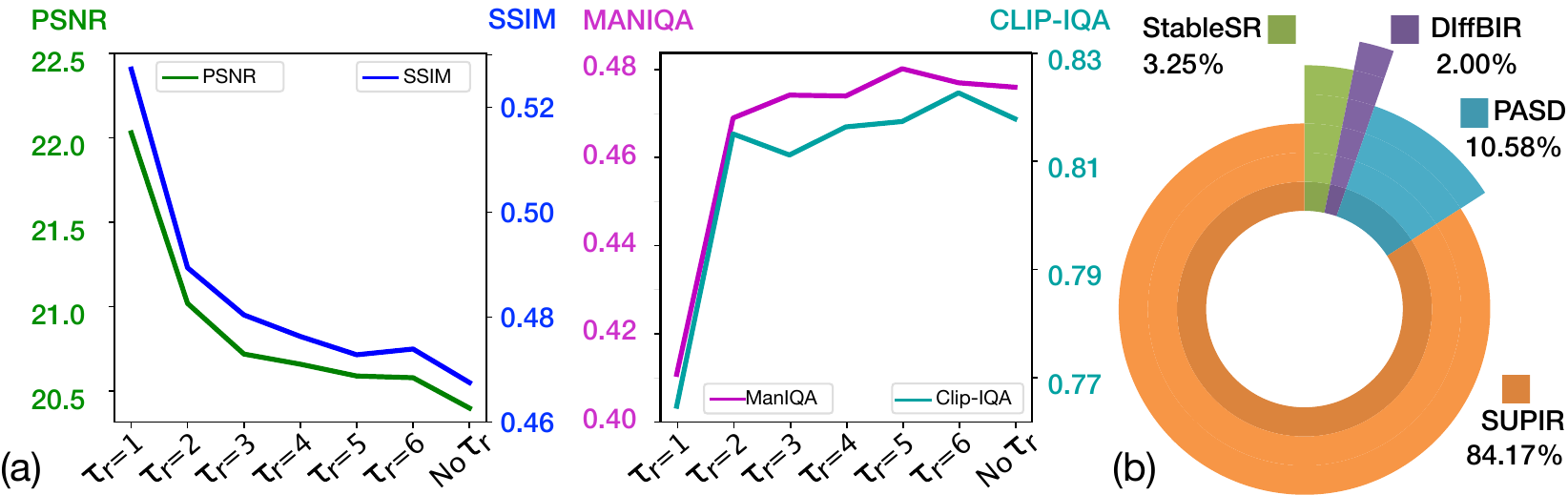}
	\vspace{-6mm}
	\caption{(a) These plots illustrate the quantitative results as a function of the variable $\tau_r$. ``No $\tau_r$'' means not to use the proposed sampling method. (b) The results of our user study.}
    \label{fig2:user_study}
	\vspace{-3mm}
\end{figure}
%%%%%%%%%%%%%%%%%%%%%%%%%%%%%%%%%%%%%%%%%%%%%%%%%%

%%%%%%%%%%%%%%%%%%%%%%%%%%%%%%%%%%%%%%%%%%%%%%%%%%

\begin{figure}[t]
	%\newlength-4mm
	%\setlength{-4mm}{-0.4cm}
	\scriptsize
	\centering
    \includegraphics[width=\linewidth]{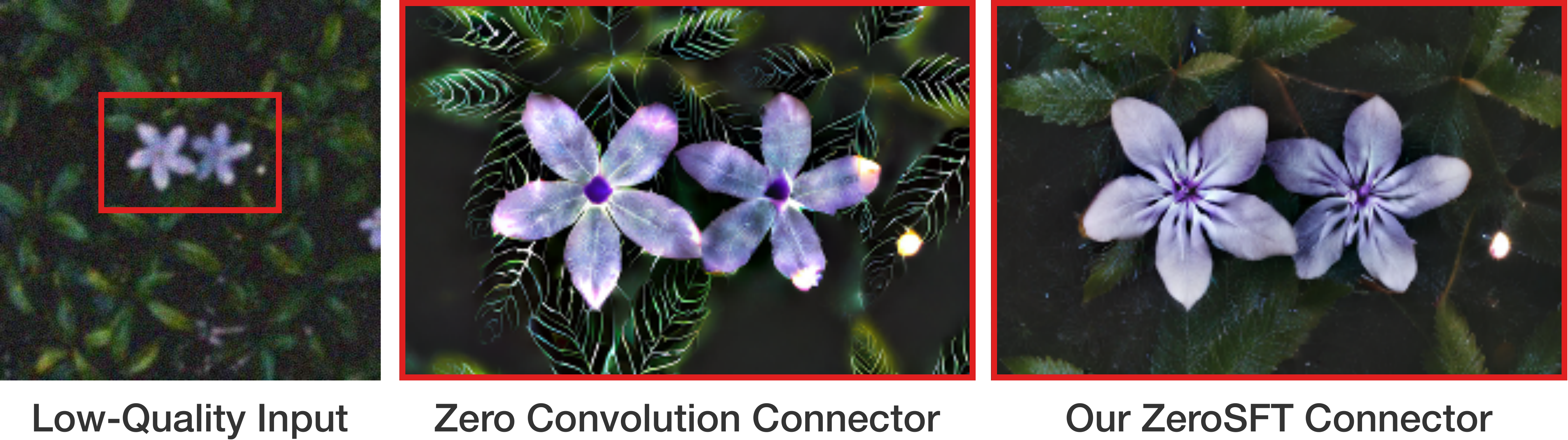}
	\vspace{-5mm}
	\caption{We compare the proposed ZeroSFT with zero convolution. Directly using zero convolution results in redundant details. The low-fidelity details can be effectively mitigated by ZeroSFT.}
    \label{fig6:projection head}
	\vspace{-4mm}
\end{figure}

%%%%%%%%%%%%%%%%%%%%%%%%%%%%%%%%%%%%%%%%%%%%%%%%%%
%%%%%%%%%%%%%%%%%%%%%%%%%%%%%%%%%%%%%%%%%%%%%%%%%%

\begin{figure*}[h]
	%\newlength-4mm
	%\setlength{-4mm}{-0.4cm}
	\scriptsize
	\centering
    \includegraphics[width=\linewidth]{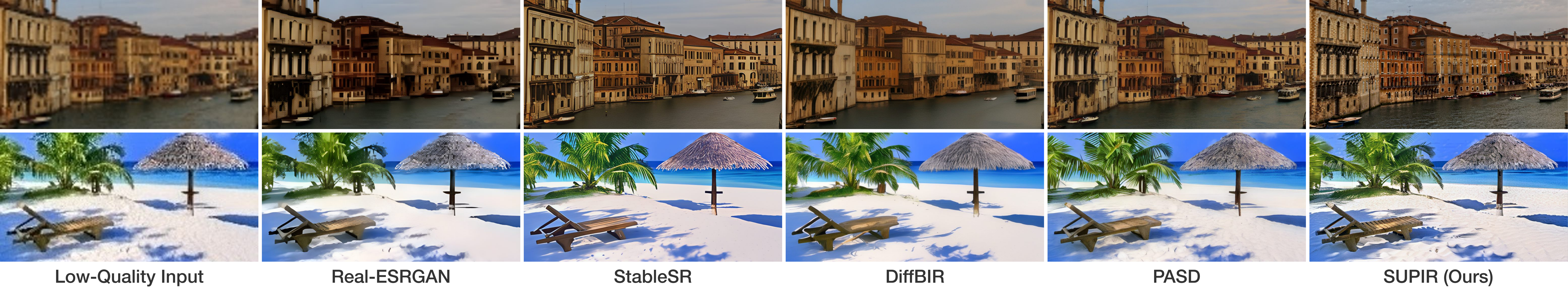}
	\vspace{-5mm}
	\caption{Qualitative comparison on real-world LQ images. \method{} successfully recovers structured buildings and lifelike rivers. It also maintains the details existing in LQ, such as the horizontal planks in the beach chairs. Zoom in for better view.}
    \label{fig:realcomp}
	\vspace{-3mm}
\end{figure*}

%%%%%%%%%%%%%%%%%%%%%%%%%%%%%%%%%%%%%%%%%%%%%%%%%%
%%%%%%%%%%%%%%%%%%%%%%%%%%%%%%%%%%%%%%%%%%%%%%%%%%

\begin{figure*}[h]
	%\newlength-4mm
	%\setlength{-4mm}{-0.4cm}
	\scriptsize
	\centering
    \includegraphics[width=\linewidth]{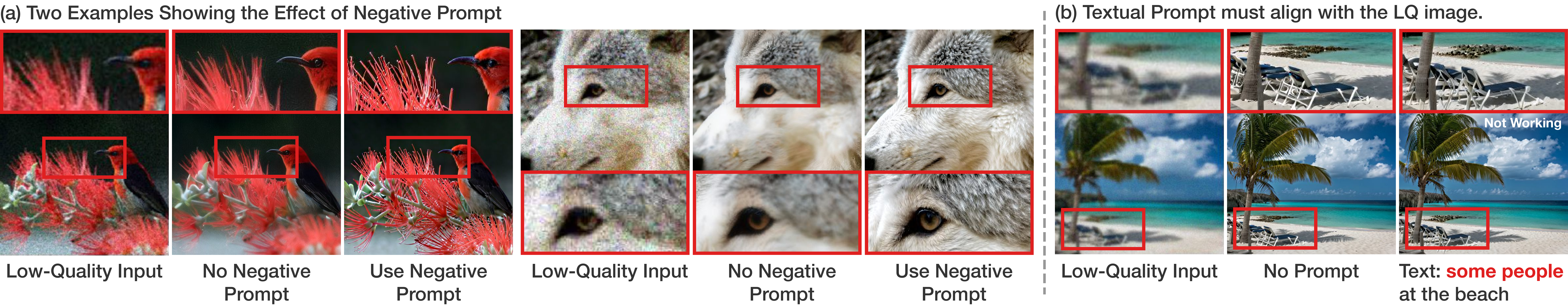}
	\vspace{-6mm}
	\caption{Influences of text prompts. (a) Negative prompts lead to detailed and sharp restoration results. 
 (b) Given a positive prompt with hallucinations, \method{} avoids generating content absent in the LQ images. Zoom in for better view.}
    \label{fig4:prompt}
	\vspace{-5mm}
\end{figure*}

%%%%%%%%%%%%%%%%%%%%%%%%%%%%%%%%%%%%%%%%%%%%%%%%%%

\vspace{-4mm}
\paragraph{Restoration in the Wild.}
We also test our method on real-world LQ images.
We collect a total of 60 real-world LQ images from RealSR \cite{cai2019toward}, DRealSR \cite{wei2020component}, Real47 \cite{lin2023diffbir}, and online sources, featuring diverse content including animals, plants, faces, buildings, and landscapes.
We show the qualitative results in \cref{fig:realcomp}, and the quantitative results are shown in \cref{tab:real}.
These results indicate that the images produced by our method have the best perceptual quality.
We also conduct a user study comparing our method on real-world LQ images, with 20 participants involved.
For each set of comparison images, we instructed participants to choose the restoration result that was of the highest quality among these test methods.
The results are shown in \cref{fig2:user_study}, revealing that our approach significantly outperformed state-of-the-art methods in perceptual quality.

\vspace{-1mm}
\subsection{Controlling Restoration with Textual Prompts}
\vspace{-2mm}
After training on a large dataset of image-text pairs and leveraging the feature of the diffusion model, our method can selectively restore images based on human prompts.
\cref{fig:teaser}(b) illustrates some examples.
In the first case, the bike restoration is challenging without prompts, but upon receiving the prompt, the model reconstructs it accurately.
In the second case, the material texture of the hat can be adjusted through prompts.
In the third case, even high-level semantic prompts allow manipulation over face attributes.
In addition to prompting the image content, we can also prompt the model to generate higher-quality images through negative-quality prompts.
\cref{fig4:prompt}(a) shows two examples.
It can be seen that the negative prompts are very effective in improving the overall quality of the output image.
We also observed that prompts in our method are not always effective.
When the provided prompts do not align with the LQ image, the prompts become ineffective, see \cref{fig4:prompt}(b).
We consider this reasonable for an IR method to stay faithful to the provided LQ image.
This reflects a significant distinction from text-to-image generation models and underscores the robustness of our approach.

\vspace{-1mm}
\subsection{Ablation Study}
\vspace{-2mm}
\paragraph{Connector.}
We compare the proposed ZeroSFT connector with zero convolution \cite{zhang2023adding}.
Quantitative results are shown in \cref{tab:connectors}.
Compared to ZeroSFT, zero convolution yields comparable performance on non-reference metrics and much lower full-reference performance.
In \cref{fig6:projection head}, we find that the drop in non-reference metrics is caused by generating low-fidelity content.
Therefore, for IR tasks, ZeroSFT ensures fidelity without losing the perceptual effect.

\begin{figure}[t]
	%\newlength-4mm
	%\setlength{-4mm}{-0.4cm}
	\scriptsize
	\centering
    \includegraphics[width=\linewidth]{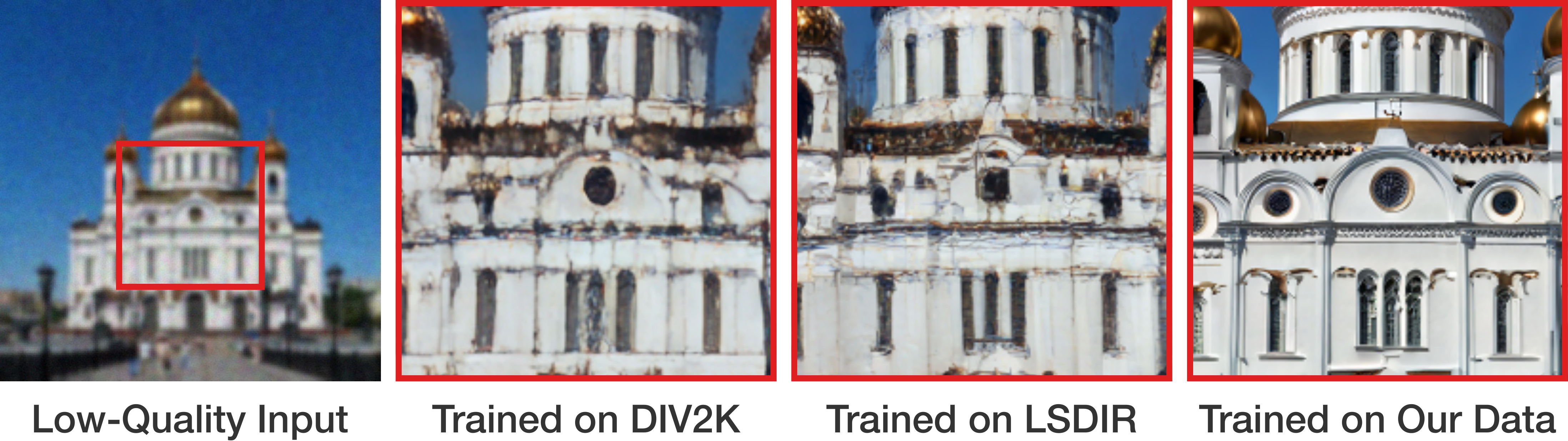}
	\vspace{-5mm}
	\caption{Qualitative comparison for \method{} training on datasets with different scales. Zoom in for better view.}
    \label{fig5:dataset}
	\vspace{-4mm}
\end{figure}

\vspace{-4mm}
\paragraph{Training data scaling.}
We trained our large-scale model on two smaller datasets for IR, DIV2K \cite{Agustsson_2017_CVPR_Workshops} and LSDIR \cite{ETHZ_Yawli}.
The qualitative results are shown in \cref{fig5:dataset}, which clearly demonstrate the importance and necessity of training on large-scale high-quality data.

%%%%%%%%%%%%%%%%%%%%%%%%%%%%%%%%%%%%%%%%%%%%%%%%%%

%%%%%%%%%%%%%%%%%%%%%%%%%%%%%%%%%%%%%%%%%%%%%%%%%%
%%%%%%%%%%%%%%%%%%%%%%%%%%%%%%%%%%%%%%%%%%%%%%%%%%

\begin{figure}[t]
	%\newlength-4mm
	%\setlength{-4mm}{-0.4cm}
	\scriptsize
	\centering
    \includegraphics[width=\linewidth]{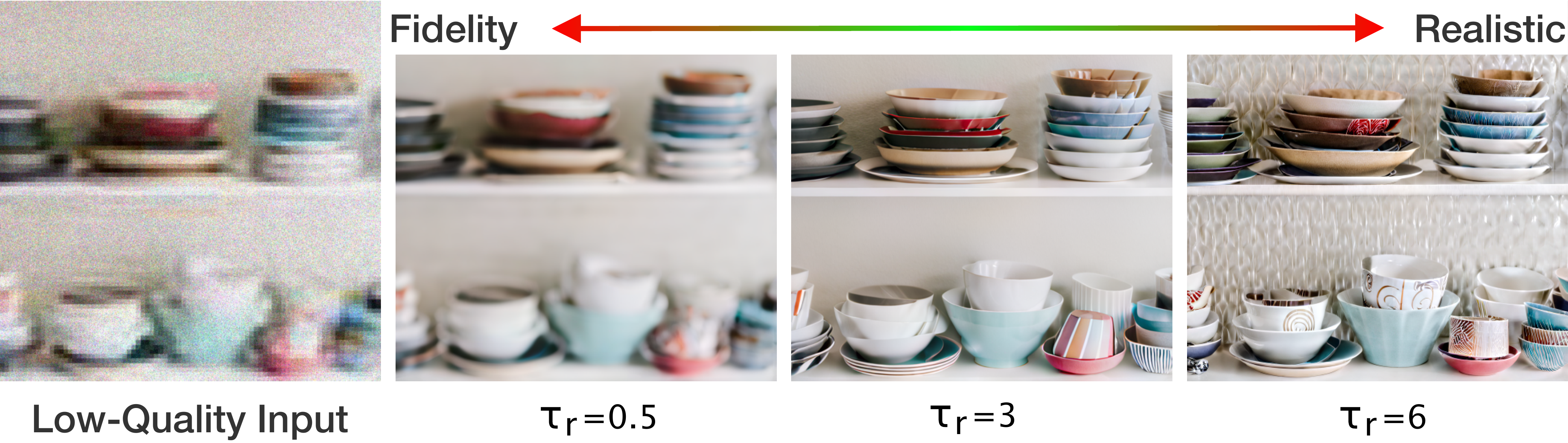}
	\vspace{-6mm}
	\caption{The effect of the proposed restoration-guided sampling method. A smaller $\tau_r$ makes the result more biased toward the LQ image, which emphasizes the fidelity. A larger $\tau_r$ emphasizes perceived quality, but with lower fidelity. Zoom in for better view.}
    \label{fig:restoration_guidance}
	\vspace{-6mm}
\end{figure}

%%%%%%%%%%%%%%%%%%%%%%%%%%%%%%%%%%%%%%%%%%%%%%%%%%

\vspace{-4mm}
\paragraph{Negative-quality samples and prompt.}
\cref{tab:prompts} shows some quantitative results under different settings.
Here, we use positive words describing image quality as ``positive prompt'', and use negative quality words and the CFG methods described in \cref{sec:3.2} as negative prompt.
It can be seen that adding positive prompts or negative prompts alone can improve the perceptual quality of the image.
Using both of them simultaneously yields the best perceptual results.
If negative samples are not included for training, these two prompts will not be able to improve the perceptual quality.
\cref{fig:neg} and \cref{fig4:prompt}(a) demonstrate the improvement in image quality brought by using negative prompts.

\vspace{-4mm}
\paragraph{Restoration-guided sampling method.}
The proposed restoration-guided sampling method is mainly controlled by $\tau_r$.
The larger $\tau_r$ is, the fewer corrections are made to the generation at each step.
The smaller $\tau_r$ is, the more generated content will be forced to be closer to the LQ image.
Please refer to \cref{fig:restoration_guidance} for a qualitative comparison.
When $\tau_r=0.5$, the image is blurry because its output is limited by the LQ image and cannot generate texture and details.
When $\tau_r=6$, there is not much guidance during generation.
The model generates a lot of texture that is not present in the LQ image, especially in flat area.
\cref{fig2:user_study}(a) illustrates the quantitative results of restoration as a function of the variable $\tau_r$.
As shown in \cref{fig2:user_study}(a), decreasing $\tau_r$ from 6 to 4 does not result in a significant decline in visual quality, while fidelity performance improves.
As restoration guidance continues to strengthen, although PSNR continues to improve, the images gradually become blurry with loss of details, as depicted in \cref{fig:restoration_guidance}.
Therefore, we choose $\tau_r=4$ as the default parameter, as it doesn't compromise image quality while effectively enhancing fidelity.

\vspace{-1mm}
\section{Conclusion}
\label{sec:conclusion}
We propose SUPIR as a pioneering IR method, empowered by model scaling, dataset enrichment, and advanced design features, expanding the horizons of IR with enhanced perceptual quality and controlled textual prompts.

\vspace{-2mm}
\paragraph{Acknowledgment.} This work was supported in part by the National Natural Science Foundation of China (62276251, 62272450), the Joint Lab of CAS-HK, the National Key R\&D Program of China (No. 2022ZD0160100), and in part by the Youth Innovation Promotion Association of Chinese Academy of Sciences (No. 2020356).

{
    \small
    \bibliographystyle{ieeenat_fullname}
    \bibliography{main}
}
% \clearpage
\section*{Appendix}
% \clearpage
\appendix

%%%%%%%%%%%%%%%%%%%%%%%%%%%%%%%%%%%%%%%%%%%%%%%%%%

\begin{figure}[t]
	%\newlength-4mm
	%\setlength{-4mm}{-0.4cm}
	\scriptsize
	\centering
    \vspace{-6mm}
    \includegraphics[width=\linewidth]{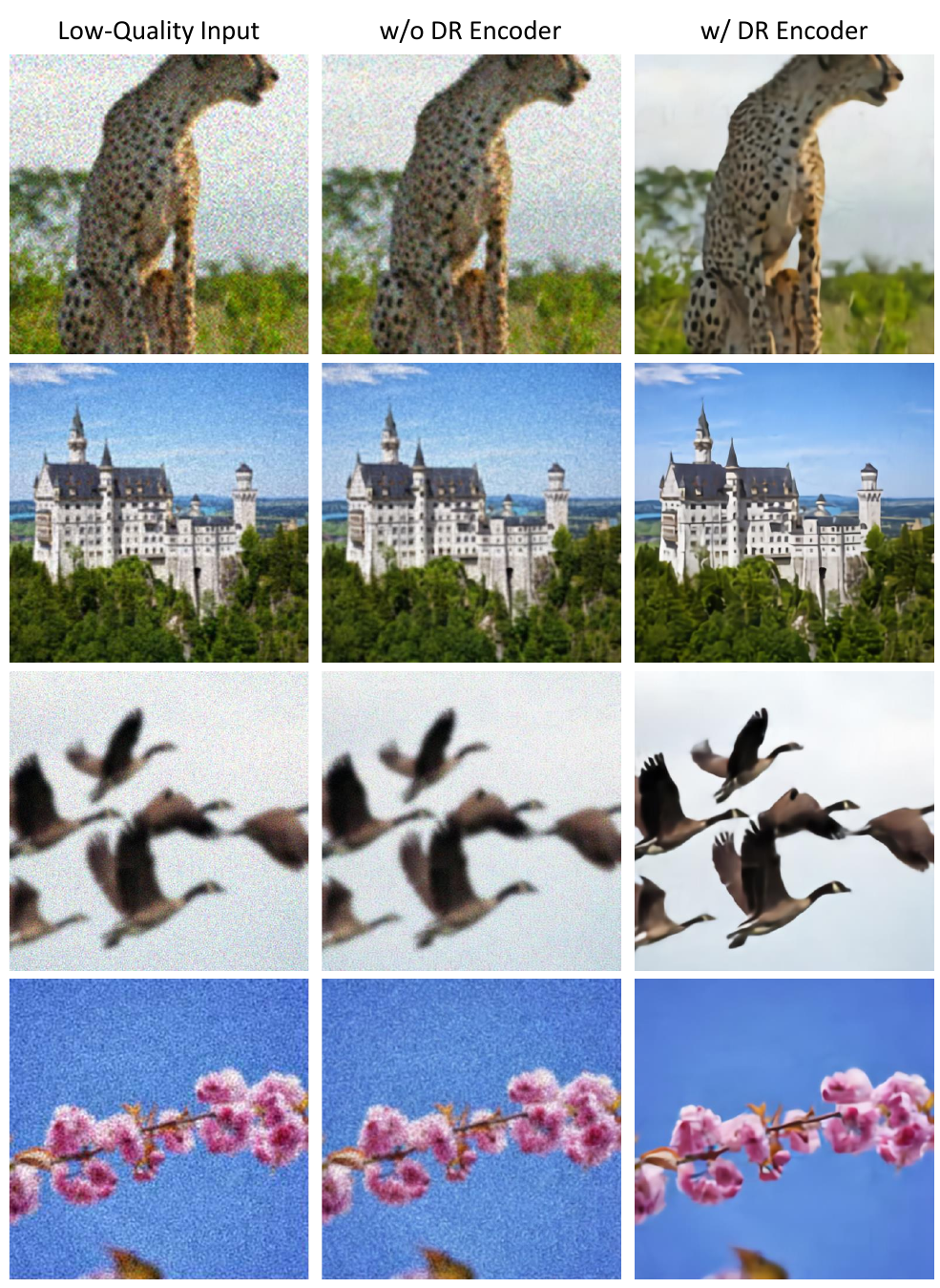}
	\vspace{-6mm}
	\caption{The effectiveness of the degradation-robust encoder (DR Encoder) is demonstrated by the results, which are achieved by initially encoding with various encoders and subsequently decoding. This process effectively reduces the degradations in low-quality inputs before they are introduced into the diffusion models.}
    \label{fig:supp_vq_denoise}
	\vspace{-5mm}
\end{figure}

%%%%%%%%%%%%%%%%%%%%%%%%%%%%%%%%%%%%%%%%%%%%%%%%%%

\section{Discussions}
\label{sec:supp_discuss}

\subsection{Degradation-Robust Encoder}
As shown in \cref{fig:pipeline} of the main text, a degradation-robust encoder is trained and deployed prior to feeding the low-quality input into the adaptor.
We conduct experiments using synthetic data to demonstrate the effectiveness of the proposed degradation-robust encoder.
In \cref{fig:supp_vq_denoise}, we show the results of using the same decoder to decode the latent representations from different encoders.
It can be seen that the original encoder has no ability to resist degradation and its decoded images still contain noise and blur.
The proposed degradation-robust encoder can reduce the impact of degradation, which further prevents generative models from misunderstanding artifacts as image content \cite{lin2023diffbir}.

\subsection{LLaVA Annotation}
Our diffusion model is capable of accepting textual prompts during the restoration process.
The prompt strategy we employ consists of two components: one component is automatically annotated by LLaVA-v1.5-13B \cite{liu2023improved}, and the other is a standardized default positive quality prompt.
The fixed portion of the prompt strategy provides a positive description of quality, including words like ``\texttt{cinematic, High Contrast, highly detailed, unreal engine, taken using a Canon EOS R camera, hyper detailed photo-realistic maximum detail, 32k, Color Grading, ultra HD, extreme meticulous detailing, skin pore detailing, hyper sharpness, perfect without deformations, Unreal Engine 5, 4k render}''.
For the LLaVA component, we use the command ``\texttt{Describe this image and its style in a very detailed manner}'' to generate detailed image captions, as exemplified in \cref{fig:supp_llava}.
While occasional inaccuracies may arise, LLaVA-v1.5-13B generally captures the essence of the low-quality input with notable precision.
Using the reconstructed version of the input proves effective in correcting these inaccuracies, allowing LLaVA to provide an accurate description of the majority of the image's content.
Additionally, \method{} is effective in mitigating the impact of potential hallucination prompts, as detailed in \cite{liu2023llava}.

%%%%%%%%%%%%%%%%%%%%%%%%%%%%%%%%%%%%%%%%%%%%%%%%%%

\begin{figure}[t]
	%\newlength-4mm
	%\setlength{-4mm}{-0.4cm}
	\scriptsize
	\centering
    \vspace{-6mm}
    \includegraphics[width=\linewidth]{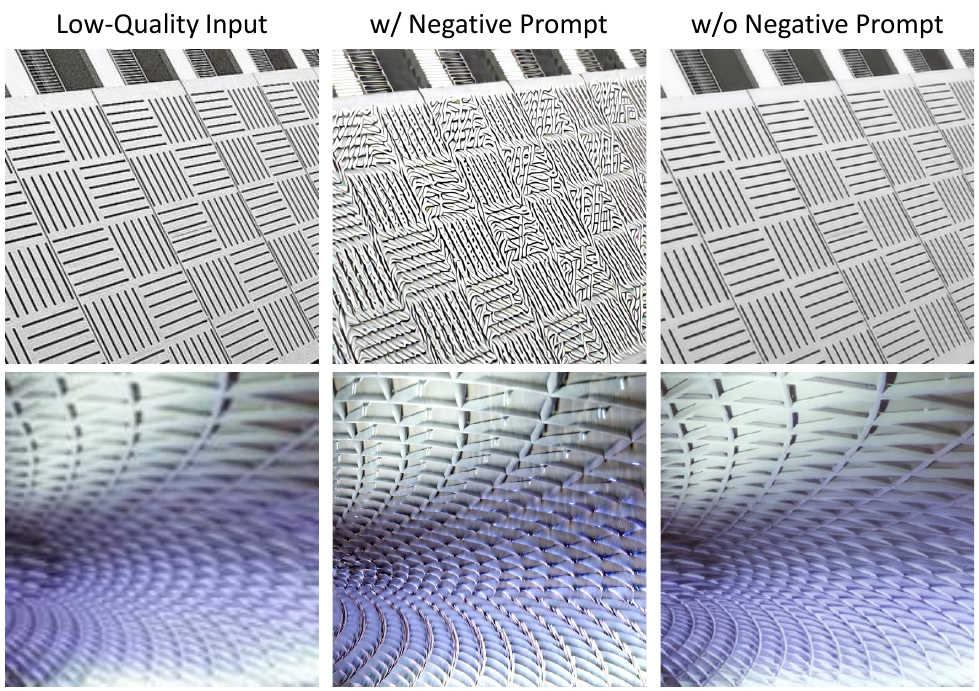}
	% \vspace{-6mm}
	\caption{Negative prompt causes artifacts when low-quality inputs do not have clear semantics.}
    \label{fig:supp_faliure}
	% \vspace{-5mm}
\end{figure}

%%%%%%%%%%%%%%%%%%%%%%%%%%%%%%%%%%%%%%%%%%%%%%%%%%

\subsection{Limitations of Negative Prompt}
\Cref{fig:supp_prompt} presents evidence that the use of negative quality prompts \cite{ho2022classifier} substantially improves the image quality of restored images.
However, as observed in \cref{fig:supp_faliure}, the negative prompt may introduce artifacts when the restoration target lacks clear semantic definition.
This issue likely stems from a misalignment between low-quality inputs and language concepts.

\subsection{Negative Samples Generation}
While negative prompts are highly effective in enhancing quality, the lack of negative-quality samples and prompts in the training data results in the fine-tuned SUPIR's inability to comprehend these prompts effectively.
To address this problem, in \cref{sec:3.2} of the main text, we introduce a method to distill negative concepts from the SDXL model.
The process for generating negative samples is illustrated in \cref{fig:supp_negative_samples}.
Direct sampling of negative samples through a text-to-image approach often results in meaningless images.
To address this issue, we also utilize training samples from our dataset as source images.
We create negative samples in an image-to-image manner as proposed in \cite{meng2021sdedit}, with a strength setting of $0.5$.

%%%%%%%%%%%%%%%%%%%%%%%%%%%%%%%%%%%%%%%%%%%%%%%%%%

\begin{figure*}[t]
	%\newlength-4mm
	%\setlength{-4mm}{-0.4cm}
	\scriptsize
	\centering
    \includegraphics[width=\linewidth]{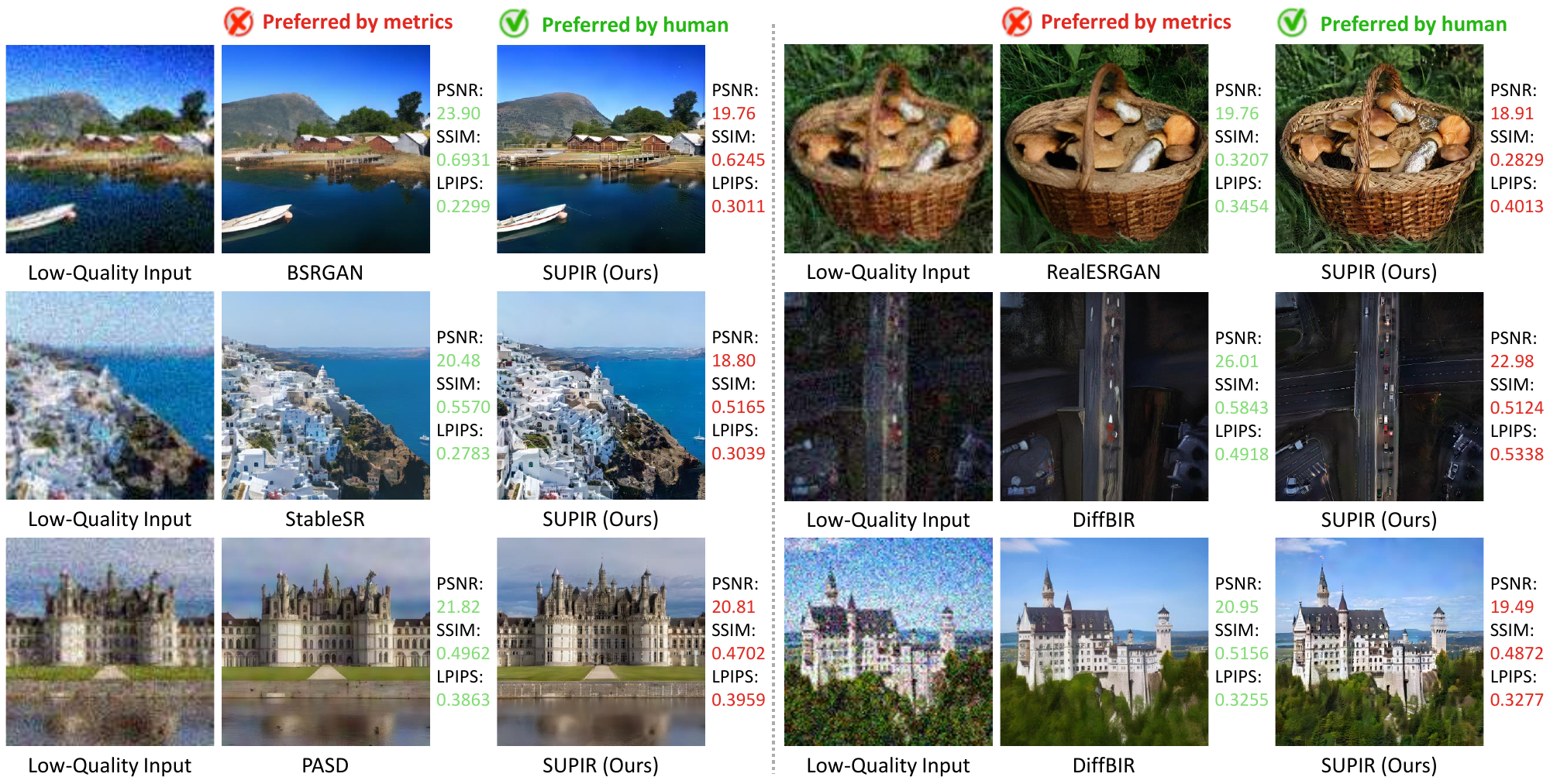}
	\vspace{-6mm}
	\caption{Additional samples highlight the misalignment between metric evaluations and human assessments. While \method{} produces images with high-fidelity textures, it tends to receive lower scores in metric evaluations.}
    \label{fig:supp_invalid_metric}
	\vspace{-5mm}
\end{figure*}

%%%%%%%%%%%%%%%%%%%%%%%%%%%%%%%%%%%%%%%%%%%%%%%%%%

%%%%%%%%%%%%%%%%%%%%%%%%%%%%%%%%%%%%%%%%%%%%%%%%%%

\begin{figure*}[t]
	%\newlength-4mm
	%\setlength{-4mm}{-0.4cm}
	\scriptsize
	\centering
    \includegraphics[width=\linewidth]{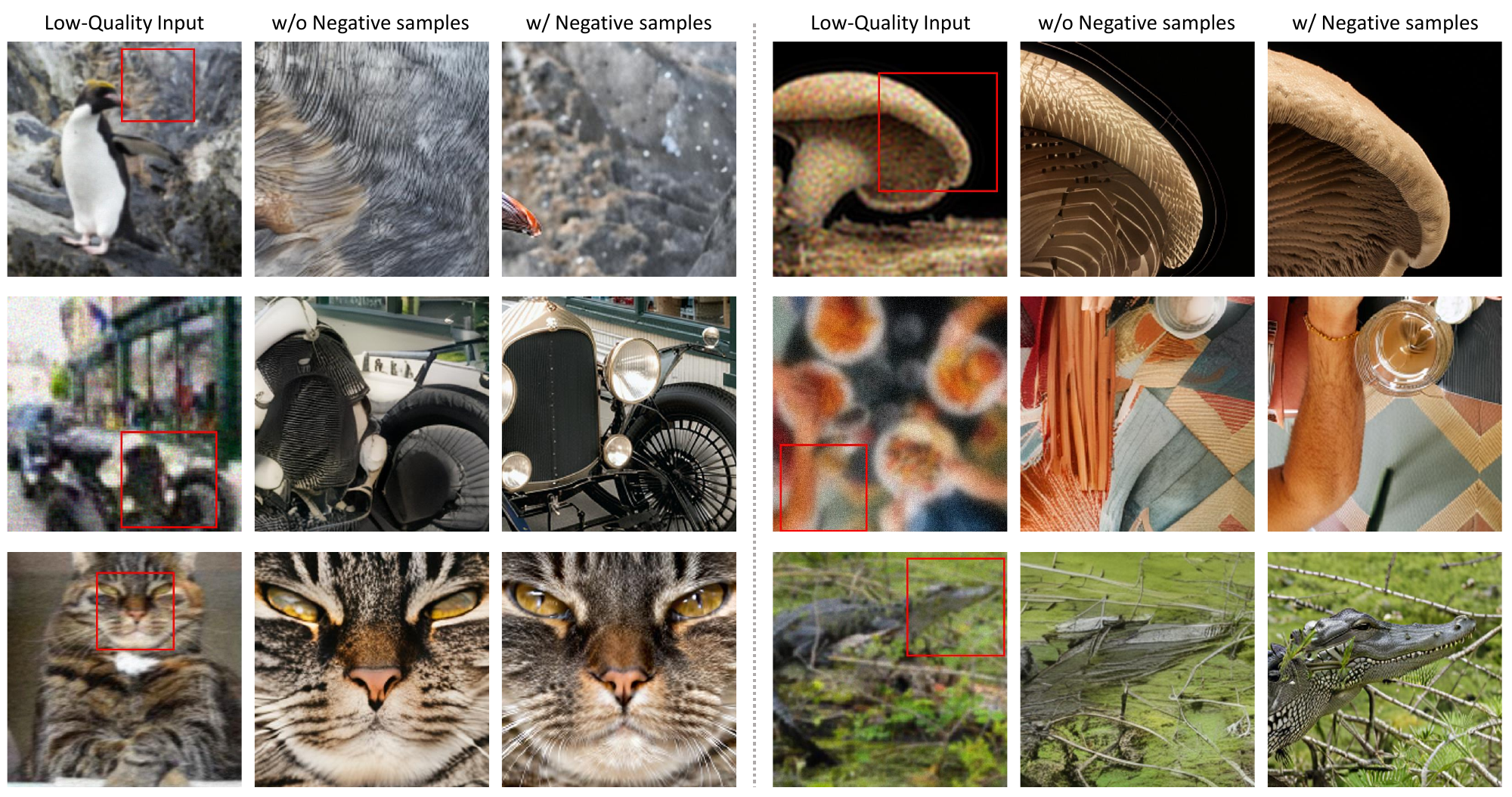}
	% \vspace{-6mm}
	\caption{More visual results for Fig. 4 of the main text. CFG introduces artifacts if we do not include negative-quality samples in training. Adding negative-quality samples allows further quality enhancement through CFG.}
    \label{fig:supp_negative_training}
	% \vspace{-5mm}
\end{figure*}

%%%%%%%%%%%%%%%%%%%%%%%%%%%%%%%%%%%%%%%%%%%%%%%%%%

\section{More Visual Results}
\label{sec:supp_more_vis}
We provide more results in this section.
\cref{fig:supp_invalid_metric}  presents additional cases where full-reference metrics do not align with human evaluation.
In \cref{fig:supp_negative_training}, we show that using negative-quality prompt without including negative samples in training may cause artifacts.
In \cref{fig:supp_visual0,fig:supp_visual1,fig:supp_visual2}, we provide more visual caparisons with other methods.
Plenty of examples prove the strong restoration ability of \method{} and the most realistic of restored images.
More examples of controllable image restoration with textual prompts can be found in \cref{fig:supp_prompt}.

%%%%%%%%%%%%%%%%%%%%%%%%%%%%%%%%%%%%%%%%%%%%%%%%%%

\begin{figure*}[t]
	%\newlength-4mm
	%\setlength{-4mm}{-0.4cm}
	\scriptsize
	\centering
    \includegraphics[width=\linewidth]{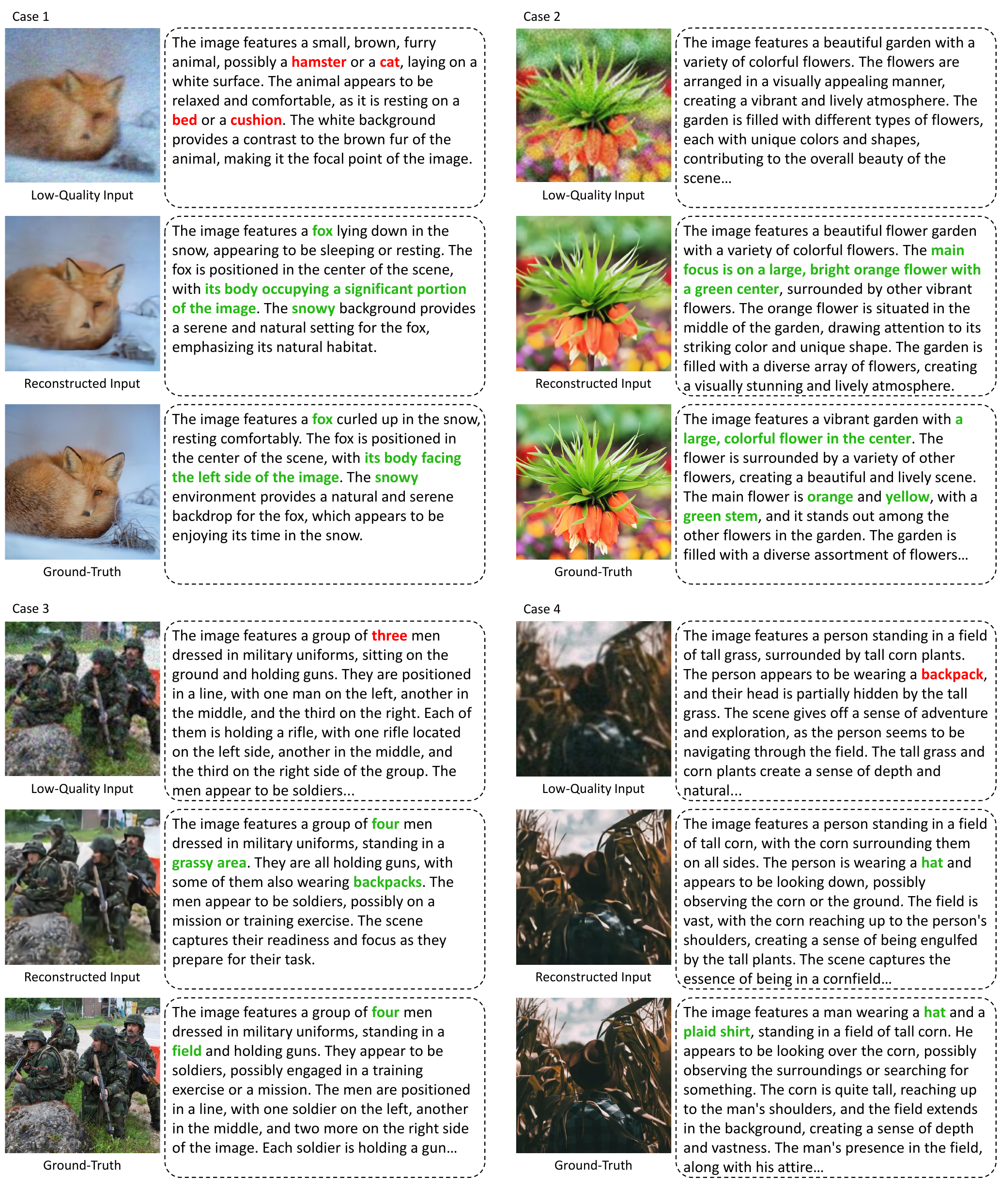}
	\vspace{-6mm}
	\caption{Snapshots showcasing LLaVA annotations demonstrate that LLaVA accurately predicts most content even with low-quality inputs. Please zoom in for a more detailed view.}
    \label{fig:supp_llava}
	\vspace{-5mm}
\end{figure*}

%%%%%%%%%%%%%%%%%%%%%%%%%%%%%%%%%%%%%%%%%%%%%%%%%%

%%%%%%%%%%%%%%%%%%%%%%%%%%%%%%%%%%%%%%%%%%%%%%%%%%

\begin{figure*}[t]
	%\newlength-4mm
	%\setlength{-4mm}{-0.4cm}
	\scriptsize
	\centering
    \includegraphics[width=\linewidth]{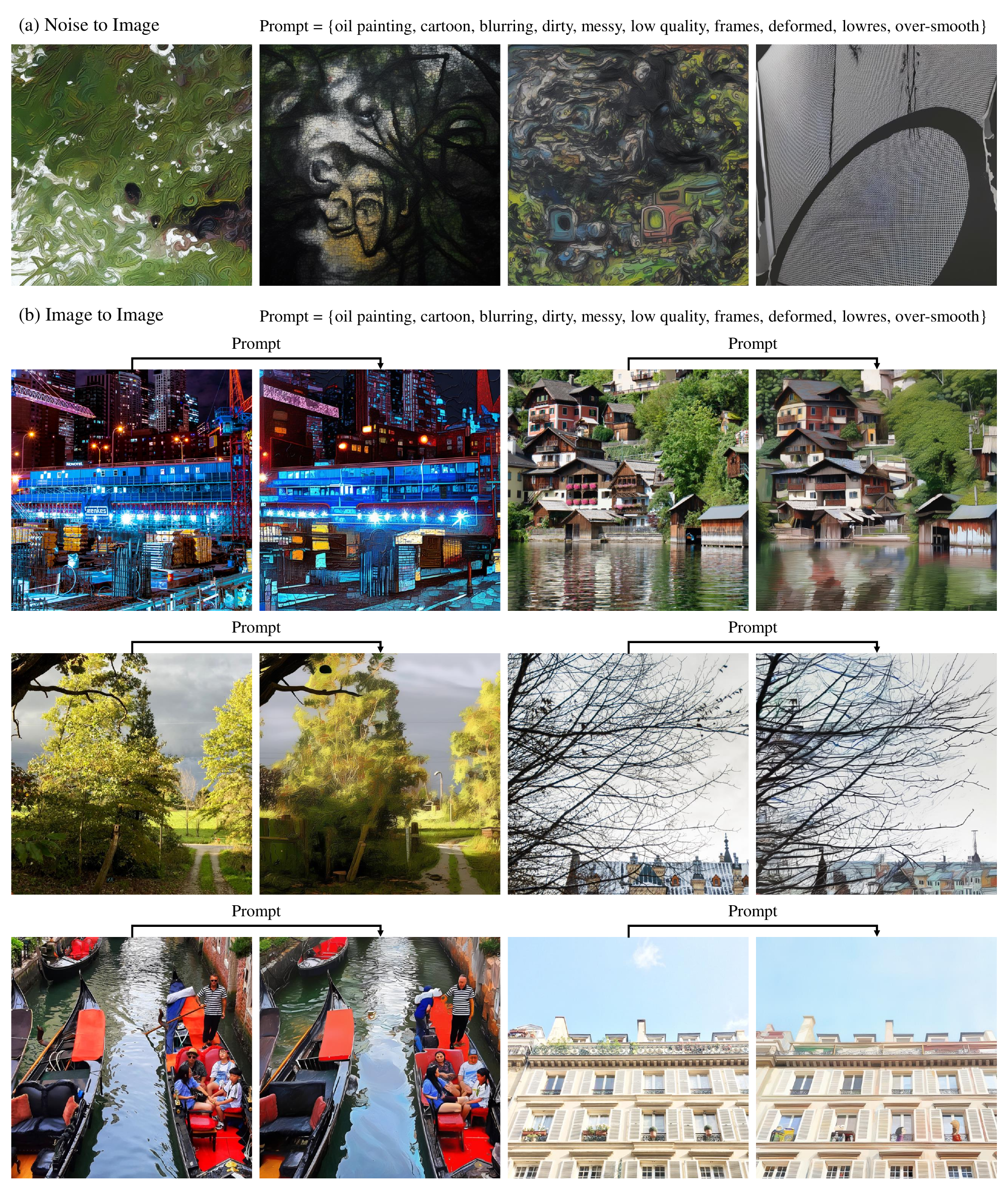}
	\vspace{-6mm}
	\caption{Pipeline of negative sample generation. (a) Sampling in a noise-to-image approach leads to meaningless outputs. (b) We synthetic negative samples from high quality images. Zoom in for better view.}
    \label{fig:supp_negative_samples}
	\vspace{-5mm}
\end{figure*}

\begin{figure*}[t]
	%\newlength-4mm
	%\setlength{-4mm}{-0.4cm}
	\scriptsize
	\centering
    \vspace{-4mm}
    \resizebox{\textwidth}{!}{
	\begin{tabular}{l}
 
		\hspace{-0.42cm}
		\begin{adjustbox}{valign=t}
			\begin{tabular}{c}
				\includegraphics[width=0.409\textwidth]{{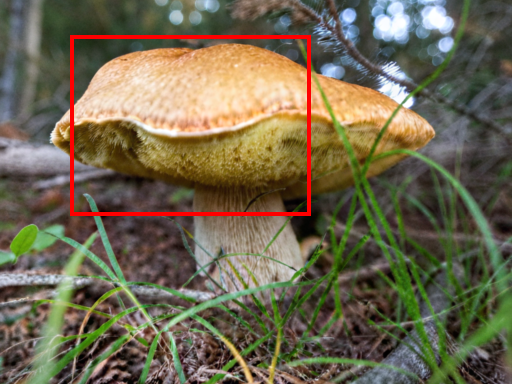}}
                \\
                \method{} (ours)
				\hspace{-10mm}
			\end{tabular}
		\end{adjustbox}
		\hspace{-5mm}
		\begin{adjustbox}{valign=t}
			\begin{tabular}{ccc}
				\includegraphics[width=0.19\textwidth]{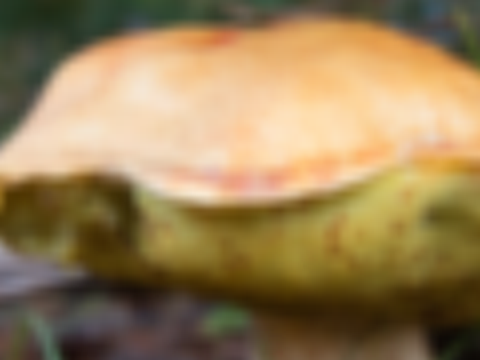} \hspace{-4.5mm} &
				\includegraphics[width=0.19\textwidth]{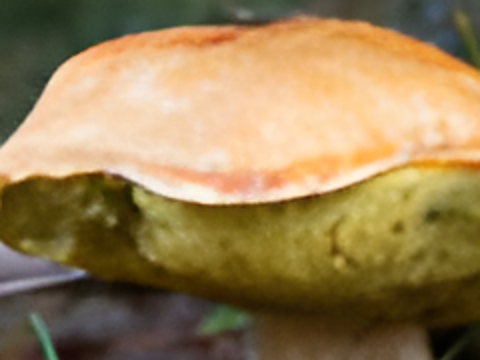} \hspace{-4.5mm} &
				\includegraphics[width=0.19\textwidth]{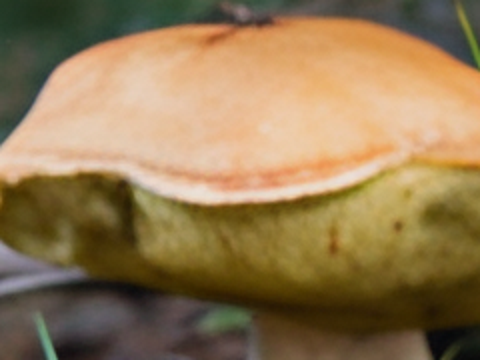} \hspace{-4.5mm} 
				\\
				Low-Quality Input \hspace{-2mm} &
				Real-ESRGAN+\cite{wang2021real} \hspace{-3.5mm} &
				StableSR\cite{wang2023exploiting} \hspace{-3.5mm}
				\\
				\includegraphics[width=0.19\textwidth]{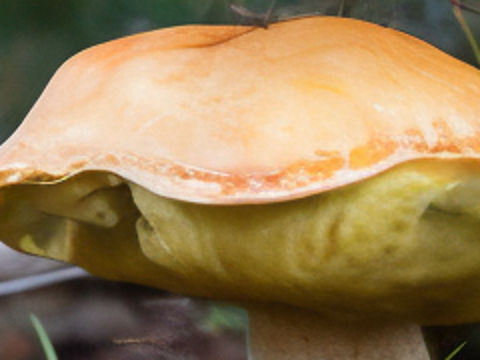} \hspace{-4.5mm} &
				\includegraphics[width=0.19\textwidth]{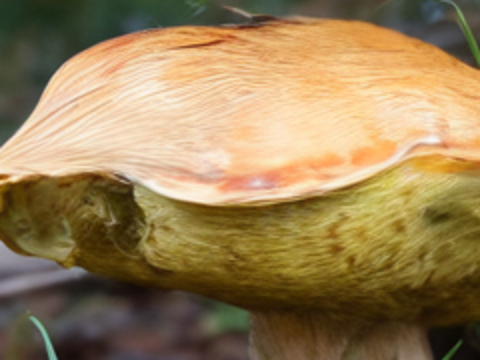} \hspace{-4.5mm} &
				\includegraphics[width=0.19\textwidth]{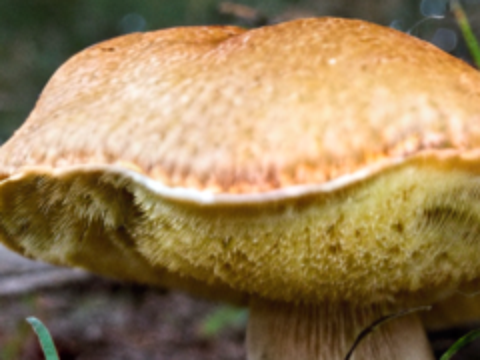} \hspace{-4.5mm}
				\\ 
				DiffBIR\cite{lin2023diffbir} \hspace{-3.5mm} &
				PASD\cite{yang2023pixel} \hspace{-3.5mm} &
				\method{} (ours) \hspace{-3.5mm}
			\end{tabular}
		\end{adjustbox}
		\hspace{-2mm}
        \\
		\hspace{-0.42cm}
		\begin{adjustbox}{valign=t}
			\begin{tabular}{c}
				\includegraphics[width=0.409\textwidth]{{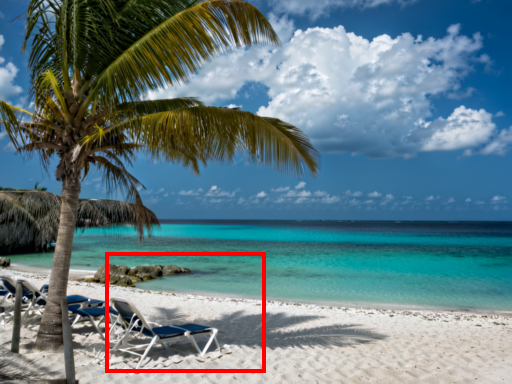}}
                \\
                \method{} (ours)
				\hspace{-10mm}
			\end{tabular}
		\end{adjustbox}
		\hspace{-5mm}
		\begin{adjustbox}{valign=t}
			\begin{tabular}{ccc}
				\includegraphics[width=0.19\textwidth]{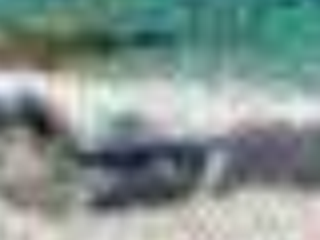} \hspace{-4.5mm} &
				\includegraphics[width=0.19\textwidth]{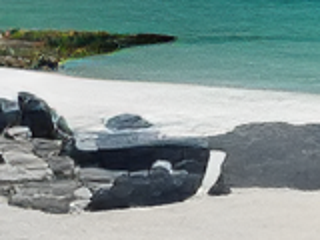} \hspace{-4.5mm} &
				\includegraphics[width=0.19\textwidth]{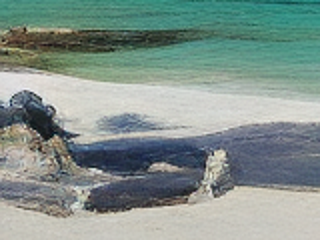} \hspace{-4.5mm} 
				\\
				Low-Quality Input \hspace{-2mm} &
				Real-ESRGAN+\cite{wang2021real} \hspace{-3.5mm} &
				StableSR\cite{wang2023exploiting} \hspace{-3.5mm}
				\\
				\includegraphics[width=0.19\textwidth]{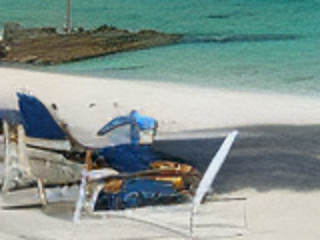} \hspace{-4.5mm} &
				\includegraphics[width=0.19\textwidth]{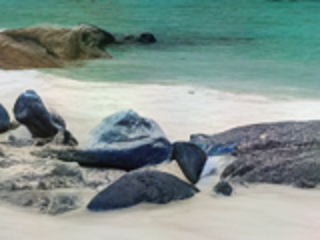} \hspace{-4.5mm} &
				\includegraphics[width=0.19\textwidth]{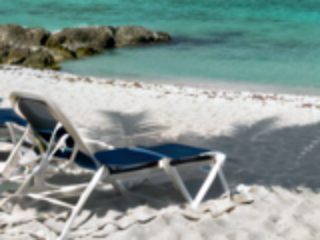} \hspace{-4.5mm}
				\\ 
				DiffBIR\cite{lin2023diffbir} \hspace{-3.5mm} &
				PASD\cite{yang2023pixel} \hspace{-3.5mm} &
				\method{} (ours) \hspace{-3.5mm}
			\end{tabular}
		\end{adjustbox}
		\hspace{-2mm}
        \\
        		\hspace{-0.42cm}
		\begin{adjustbox}{valign=t}
			\begin{tabular}{c}
				\includegraphics[width=0.409\textwidth]{{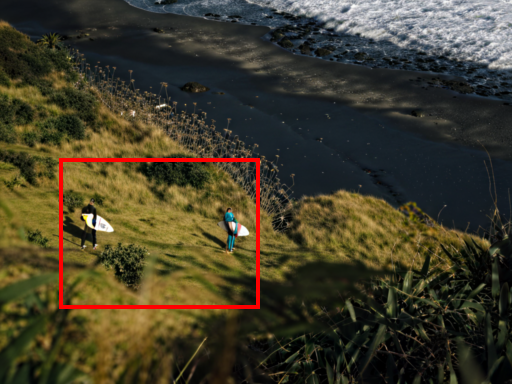}}
                \\
                \method{} (ours)
				\hspace{-10mm}
			\end{tabular}
		\end{adjustbox}
		\hspace{-5mm}
		\begin{adjustbox}{valign=t}
			\begin{tabular}{ccc}
				\includegraphics[width=0.19\textwidth]{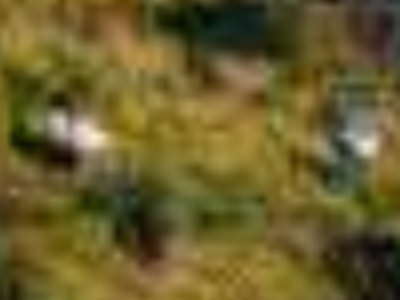} \hspace{-4.5mm} &
				\includegraphics[width=0.19\textwidth]{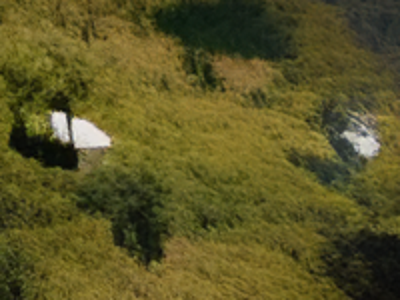} \hspace{-4.5mm} &
				\includegraphics[width=0.19\textwidth]{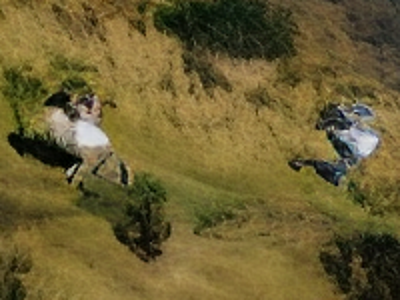} \hspace{-4.5mm} 
				\\
				Low-Quality Input \hspace{-2mm} &
				Real-ESRGAN+\cite{wang2021real} \hspace{-3.5mm} &
				StableSR\cite{wang2023exploiting} \hspace{-3.5mm}
				\\
				\includegraphics[width=0.19\textwidth]{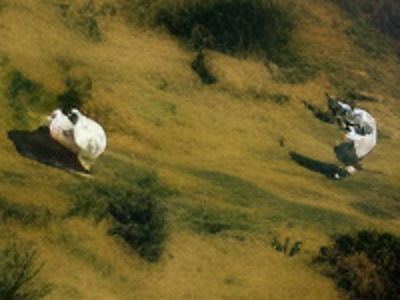} \hspace{-4.5mm} &
				\includegraphics[width=0.19\textwidth]{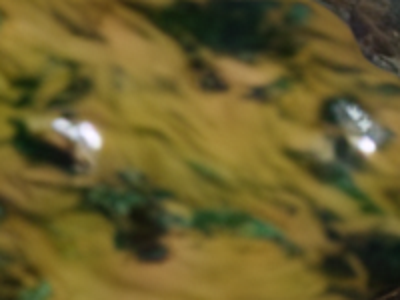} \hspace{-4.5mm} &
				\includegraphics[width=0.19\textwidth]{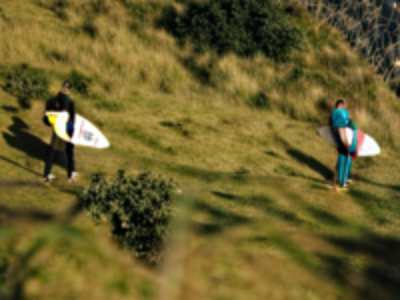} \hspace{-4.5mm}
				\\ 
				DiffBIR\cite{lin2023diffbir} \hspace{-3.5mm} &
				PASD\cite{yang2023pixel} \hspace{-3.5mm} &
				\method{} (ours) \hspace{-3.5mm}
			\end{tabular}
		\end{adjustbox}
		\hspace{-2mm}
        \\
        		\hspace{-0.42cm}
		\begin{adjustbox}{valign=t}
			\begin{tabular}{c}
				\includegraphics[width=0.409\textwidth]{{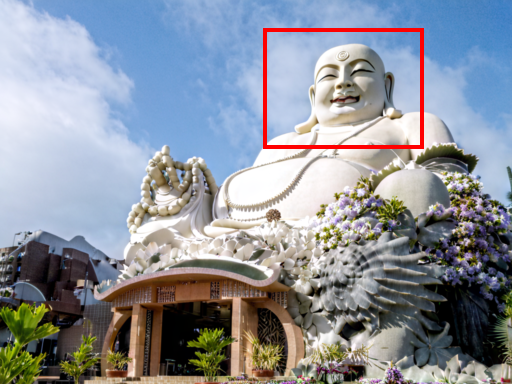}}
                \\
                \method{} (ours)
				\hspace{-10mm}
			\end{tabular}
		\end{adjustbox}
		\hspace{-5mm}
		\begin{adjustbox}{valign=t}
			\begin{tabular}{ccc}
				\includegraphics[width=0.19\textwidth]{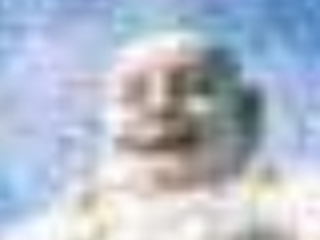} \hspace{-4.5mm} &
				\includegraphics[width=0.19\textwidth]{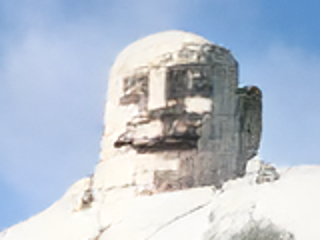} \hspace{-4.5mm} &
				\includegraphics[width=0.19\textwidth]{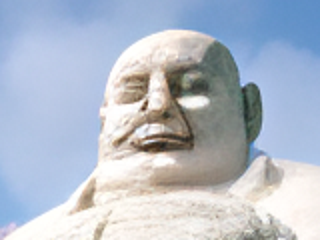} \hspace{-4.5mm} 
				\\
				Low-Quality Input \hspace{-2mm} &
				Real-ESRGAN+\cite{wang2021real} \hspace{-3.5mm} &
				StableSR\cite{wang2023exploiting} \hspace{-3.5mm}
				\\
				\includegraphics[width=0.19\textwidth]{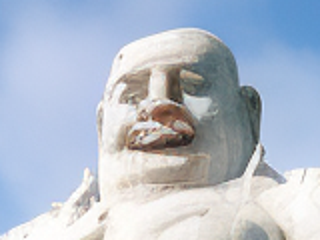} \hspace{-4.5mm} &
				\includegraphics[width=0.19\textwidth]{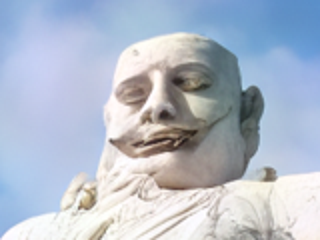} \hspace{-4.5mm} &
				\includegraphics[width=0.19\textwidth]{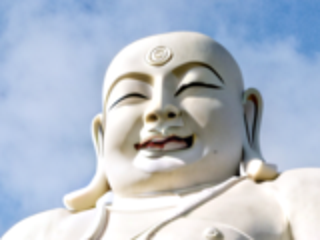} \hspace{-4.5mm}
				\\ 
				DiffBIR\cite{lin2023diffbir} \hspace{-3.5mm} &
				PASD\cite{yang2023pixel} \hspace{-3.5mm} &
				\method{} (ours) \hspace{-3.5mm}
			\end{tabular}
		\end{adjustbox}
	\end{tabular}
    }
    % \hspace{-8mm}
    \vspace{-4mm}
	\caption{Qualitative comparison with different methods. Our method can accurately restore the texture and details of the corresponding object under challenging degradation. Zoom in for better view.}
    \label{fig:supp_visual0}
	\vspace{-6mm}
\end{figure*}

\begin{figure*}[t]
	%\newlength-4mm
	%\setlength{-4mm}{-0.4cm}
	\scriptsize
	\centering
    \vspace{-4mm}
    \resizebox{\textwidth}{!}{
	\begin{tabular}{l}
 
		\hspace{-0.42cm}
		\begin{adjustbox}{valign=t}
			\begin{tabular}{c}
				\includegraphics[width=0.409\textwidth]{{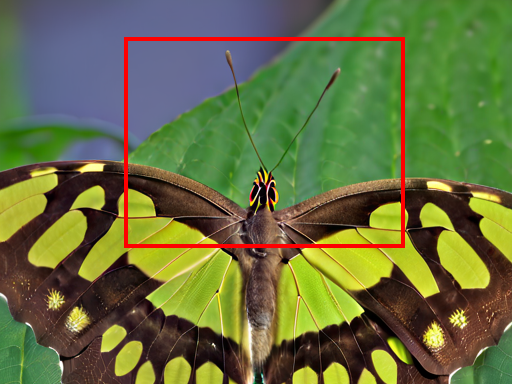}}
                \\
                \method{} (ours)
				\hspace{-10mm}
			\end{tabular}
		\end{adjustbox}
		\hspace{-5mm}
		\begin{adjustbox}{valign=t}
			\begin{tabular}{ccc}
				\includegraphics[width=0.19\textwidth]{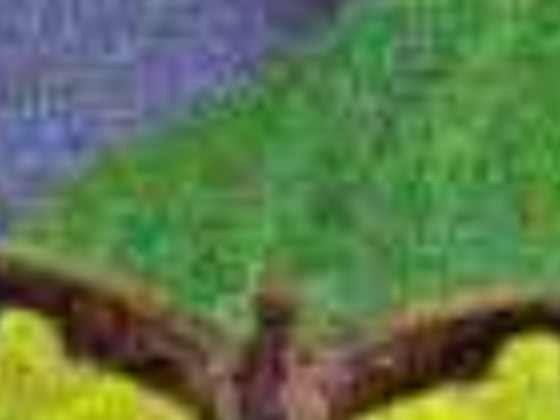} \hspace{-4.5mm} &
				\includegraphics[width=0.19\textwidth]{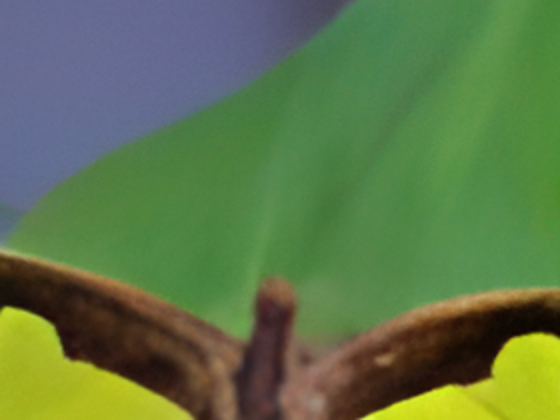} \hspace{-4.5mm} &
				\includegraphics[width=0.19\textwidth]{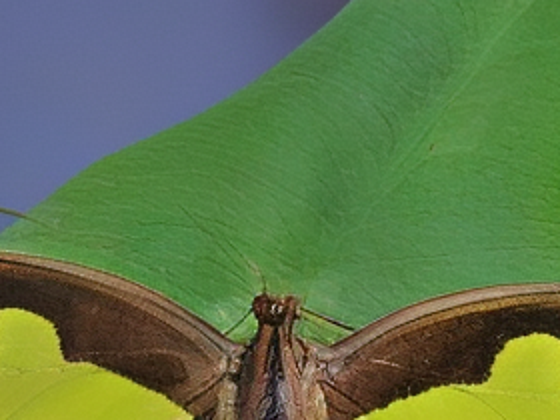} \hspace{-4.5mm} 
				\\
				Low-Quality Input \hspace{-2mm} &
				Real-ESRGAN+\cite{wang2021real} \hspace{-3.5mm} &
				StableSR\cite{wang2023exploiting} \hspace{-3.5mm}
				\\
				\includegraphics[width=0.19\textwidth]{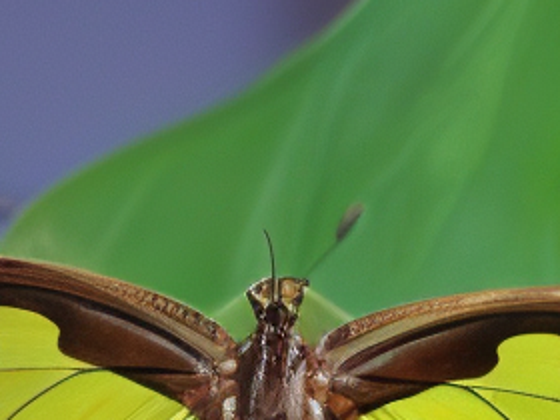} \hspace{-4.5mm} &
				\includegraphics[width=0.19\textwidth]{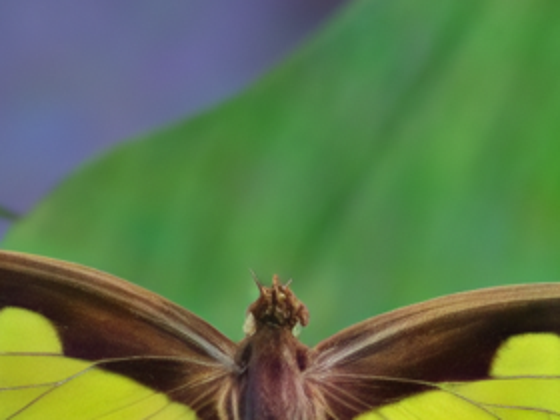} \hspace{-4.5mm} &
				\includegraphics[width=0.19\textwidth]{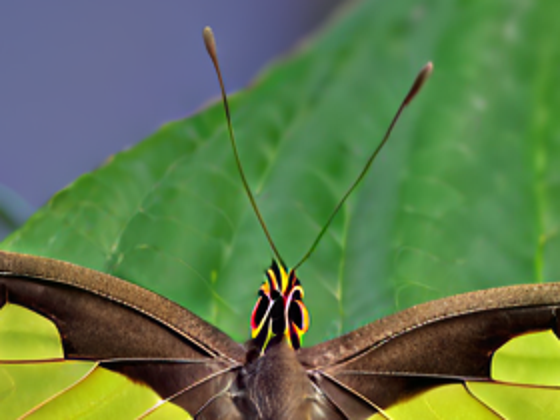} \hspace{-4.5mm}
				\\ 
				DiffBIR\cite{lin2023diffbir} \hspace{-3.5mm} &
				PASD\cite{yang2023pixel} \hspace{-3.5mm} &
				\method{} (ours) \hspace{-3.5mm}
			\end{tabular}
		\end{adjustbox}
		\hspace{-2mm}
        \\
		\hspace{-0.42cm}
		\begin{adjustbox}{valign=t}
			\begin{tabular}{c}
				\includegraphics[width=0.409\textwidth]{{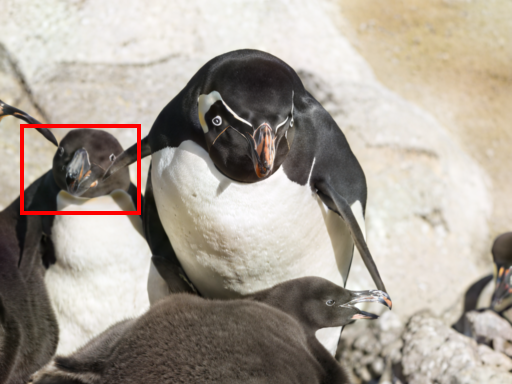}}
                \\
                \method{} (ours)
				\hspace{-10mm}
			\end{tabular}
		\end{adjustbox}
		\hspace{-5mm}
		\begin{adjustbox}{valign=t}
			\begin{tabular}{ccc}
				\includegraphics[width=0.19\textwidth]{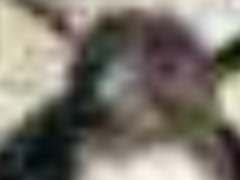} \hspace{-4.5mm} &
				\includegraphics[width=0.19\textwidth]{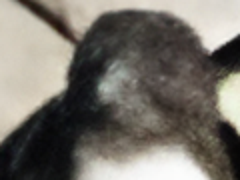} \hspace{-4.5mm} &
				\includegraphics[width=0.19\textwidth]{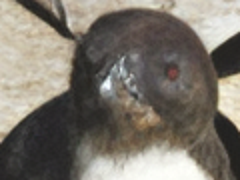} \hspace{-4.5mm} 
				\\
				Low-Quality Input \hspace{-2mm} &
				Real-ESRGAN+\cite{wang2021real} \hspace{-3.5mm} &
				StableSR\cite{wang2023exploiting} \hspace{-3.5mm}
				\\
				\includegraphics[width=0.19\textwidth]{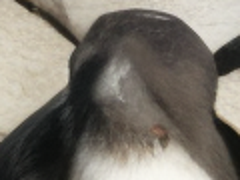} \hspace{-4.5mm} &
				\includegraphics[width=0.19\textwidth]{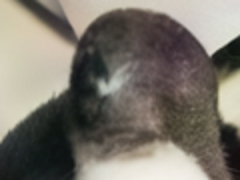} \hspace{-4.5mm} &
				\includegraphics[width=0.19\textwidth]{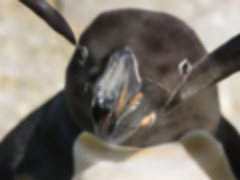} \hspace{-4.5mm}
				\\ 
				DiffBIR\cite{lin2023diffbir} \hspace{-3.5mm} &
				PASD\cite{yang2023pixel} \hspace{-3.5mm} &
				\method{} (ours) \hspace{-3.5mm}
			\end{tabular}
		\end{adjustbox}
		\hspace{-2mm}
        \\
        		\hspace{-0.42cm}
		\begin{adjustbox}{valign=t}
			\begin{tabular}{c}
				\includegraphics[width=0.409\textwidth]{{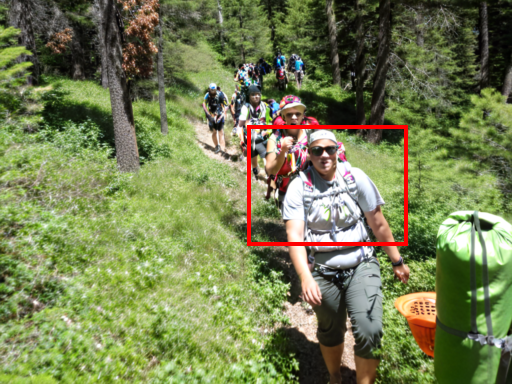}}
                \\
                \method{} (ours)
				\hspace{-10mm}
			\end{tabular}
		\end{adjustbox}
		\hspace{-5mm}
		\begin{adjustbox}{valign=t}
			\begin{tabular}{ccc}
				\includegraphics[width=0.19\textwidth]{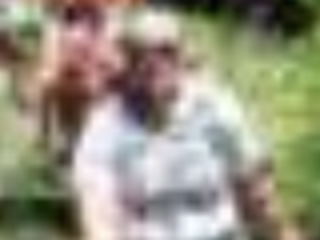} \hspace{-4.5mm} &
				\includegraphics[width=0.19\textwidth]{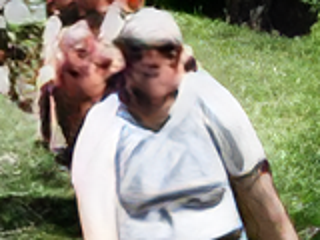} \hspace{-4.5mm} &
				\includegraphics[width=0.19\textwidth]{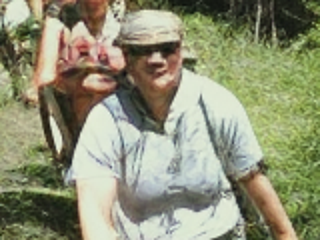} \hspace{-4.5mm} 
				\\
				Low-Quality Input \hspace{-2mm} &
				Real-ESRGAN+\cite{wang2021real} \hspace{-3.5mm} &
				StableSR\cite{wang2023exploiting} \hspace{-3.5mm}
				\\
				\includegraphics[width=0.19\textwidth]{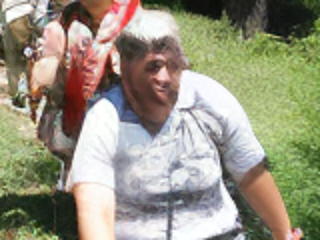} \hspace{-4.5mm} &
				\includegraphics[width=0.19\textwidth]{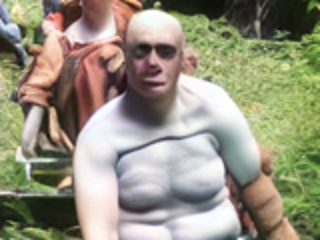} \hspace{-4.5mm} &
				\includegraphics[width=0.19\textwidth]{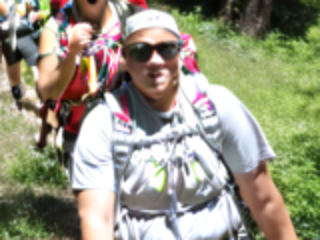} \hspace{-4.5mm}
				\\ 
				DiffBIR\cite{lin2023diffbir} \hspace{-3.5mm} &
				PASD\cite{yang2023pixel} \hspace{-3.5mm} &
				\method{} (ours) \hspace{-3.5mm}
			\end{tabular}
		\end{adjustbox}
		\hspace{-2mm}
        \\
        		\hspace{-0.42cm}
		\begin{adjustbox}{valign=t}
			\begin{tabular}{c}
				\includegraphics[width=0.409\textwidth]{{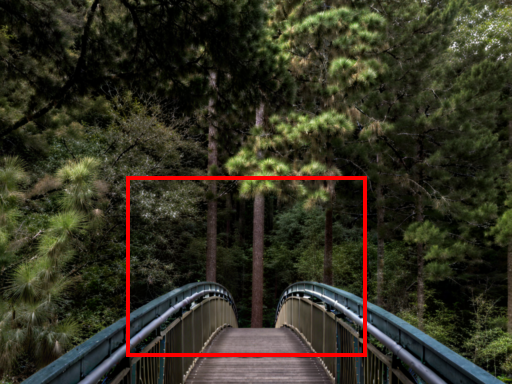}}
                \\
                \method{} (ours)
				\hspace{-10mm}
			\end{tabular}
		\end{adjustbox}
		\hspace{-5mm}
		\begin{adjustbox}{valign=t}
			\begin{tabular}{ccc}
				\includegraphics[width=0.19\textwidth]{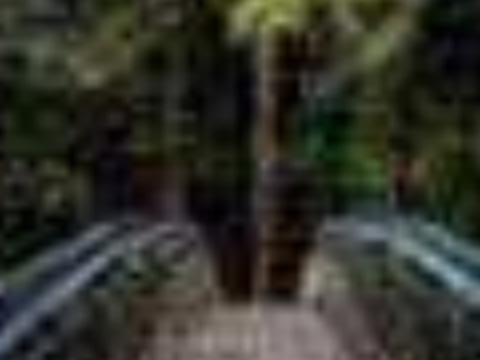} \hspace{-4.5mm} &
				\includegraphics[width=0.19\textwidth]{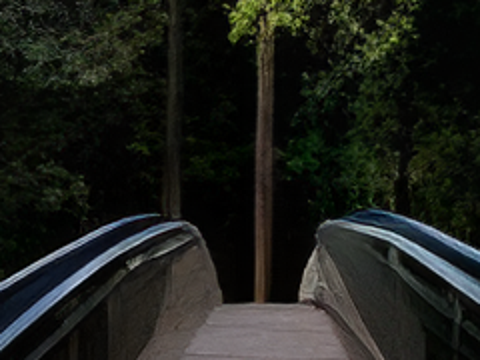} \hspace{-4.5mm} &
				\includegraphics[width=0.19\textwidth]{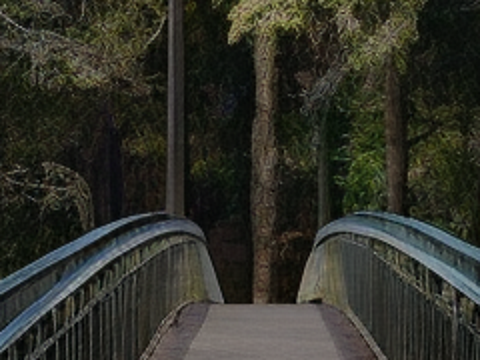} \hspace{-4.5mm} 
				\\
				Low-Quality Input \hspace{-2mm} &
				Real-ESRGAN+\cite{wang2021real} \hspace{-3.5mm} &
				StableSR\cite{wang2023exploiting} \hspace{-3.5mm}
				\\
				\includegraphics[width=0.19\textwidth]{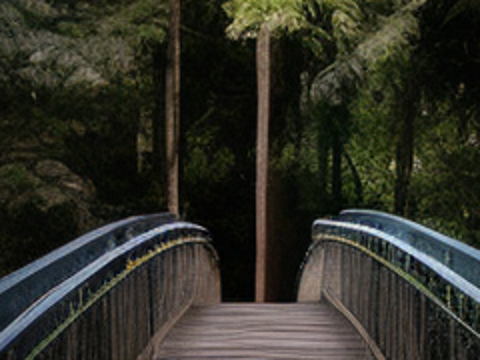} \hspace{-4.5mm} &
				\includegraphics[width=0.19\textwidth]{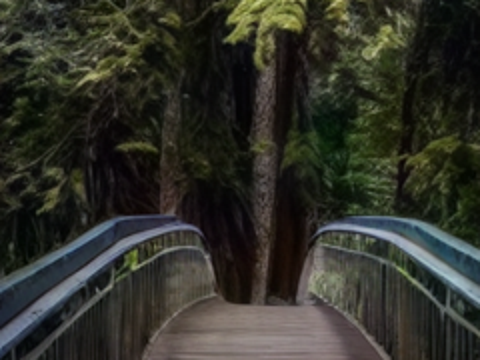} \hspace{-4.5mm} &
				\includegraphics[width=0.19\textwidth]{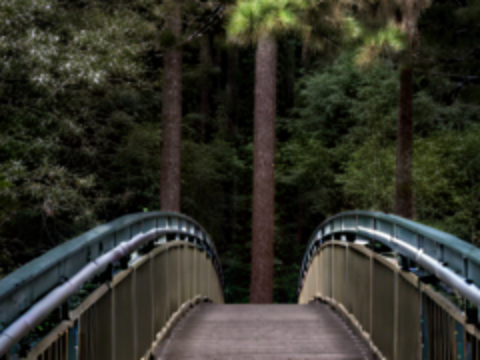} \hspace{-4.5mm}
				\\ 
				DiffBIR\cite{lin2023diffbir} \hspace{-3.5mm} &
				PASD\cite{yang2023pixel} \hspace{-3.5mm} &
				\method{} (ours) \hspace{-3.5mm}
			\end{tabular}
		\end{adjustbox}
	\end{tabular}
    }
    % \hspace{-8mm}
    \vspace{-4mm}
	\caption{Qualitative comparison with different methods. Our method can accurately restore the texture and details of the corresponding object under challenging degradation. Zoom in for better view.}
    \label{fig:supp_visual1}
	\vspace{-6mm}
\end{figure*}

\begin{figure*}[t]
	%\newlength-4mm
	%\setlength{-4mm}{-0.4cm}
	\scriptsize
	\centering
    \vspace{-4mm}
    \resizebox{\textwidth}{!}{
	\begin{tabular}{l}
 
		\hspace{-0.42cm}
		\begin{adjustbox}{valign=t}
			\begin{tabular}{c}
				\includegraphics[width=0.409\textwidth]{{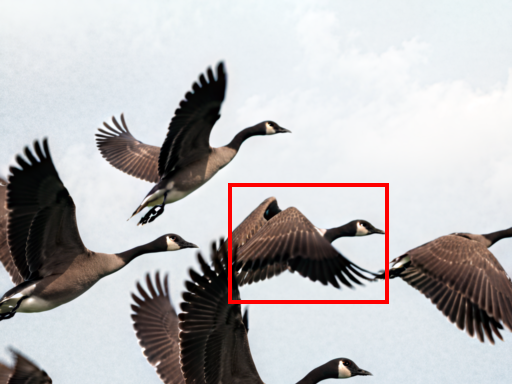}}
                \\
                \method{} (ours)
				\hspace{-10mm}
			\end{tabular}
		\end{adjustbox}
		\hspace{-5mm}
		\begin{adjustbox}{valign=t}
			\begin{tabular}{ccc}
				\includegraphics[width=0.19\textwidth]{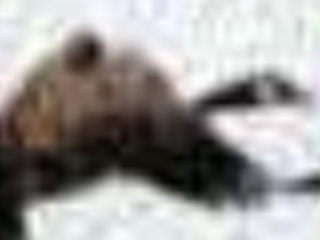} \hspace{-4.5mm} &
				\includegraphics[width=0.19\textwidth]{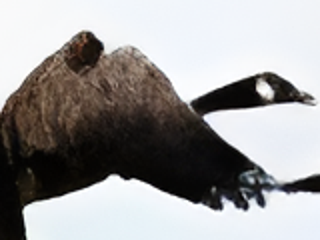} \hspace{-4.5mm} &
				\includegraphics[width=0.19\textwidth]{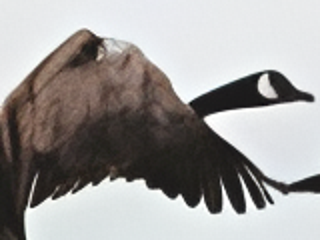} \hspace{-4.5mm} 
				\\
				Low-Quality Input \hspace{-2mm} &
				Real-ESRGAN+\cite{wang2021real} \hspace{-3.5mm} &
				StableSR\cite{wang2023exploiting} \hspace{-3.5mm}
				\\
				\includegraphics[width=0.19\textwidth]{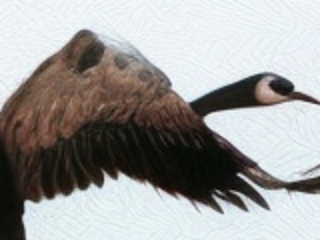} \hspace{-4.5mm} &
				\includegraphics[width=0.19\textwidth]{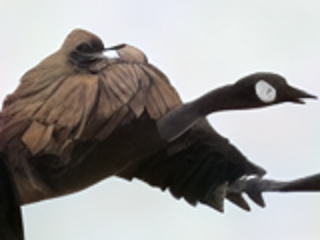} \hspace{-4.5mm} &
				\includegraphics[width=0.19\textwidth]{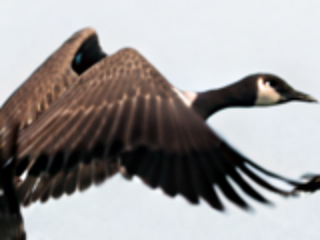} \hspace{-4.5mm}
				\\ 
				DiffBIR\cite{lin2023diffbir} \hspace{-3.5mm} &
				PASD\cite{yang2023pixel} \hspace{-3.5mm} &
				\method{} (ours) \hspace{-3.5mm}
			\end{tabular}
		\end{adjustbox}
		\hspace{-2mm}
        \\
		\hspace{-0.42cm}
		\begin{adjustbox}{valign=t}
			\begin{tabular}{c}
				\includegraphics[width=0.409\textwidth]{{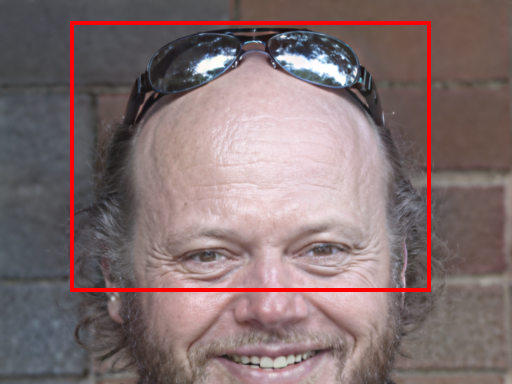}}
                \\
                \method{} (ours)
				\hspace{-10mm}
			\end{tabular}
		\end{adjustbox}
		\hspace{-5mm}
		\begin{adjustbox}{valign=t}
			\begin{tabular}{ccc}
				\includegraphics[width=0.19\textwidth]{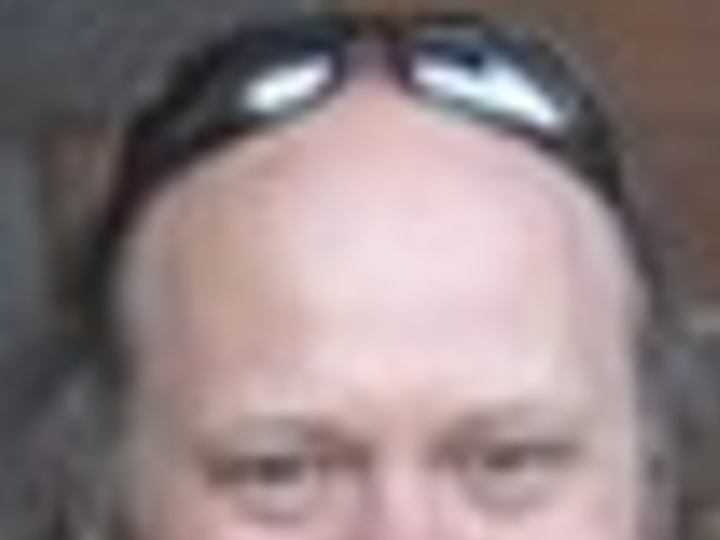} \hspace{-4.5mm} &
				\includegraphics[width=0.19\textwidth]{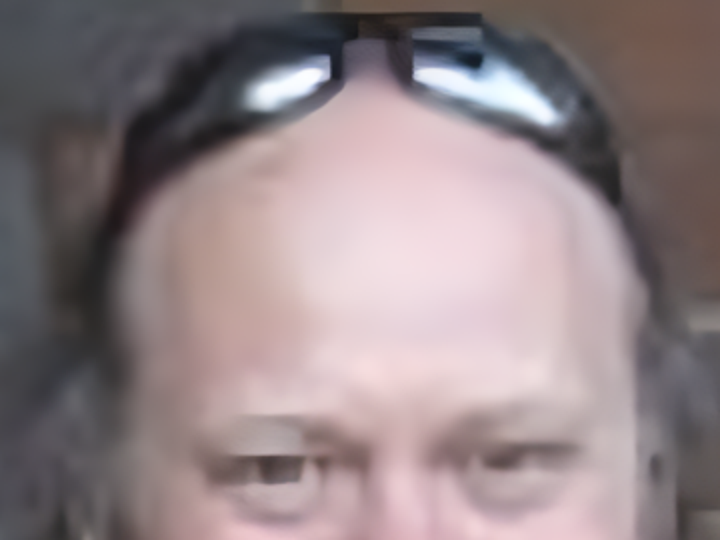} \hspace{-4.5mm} &
				\includegraphics[width=0.19\textwidth]{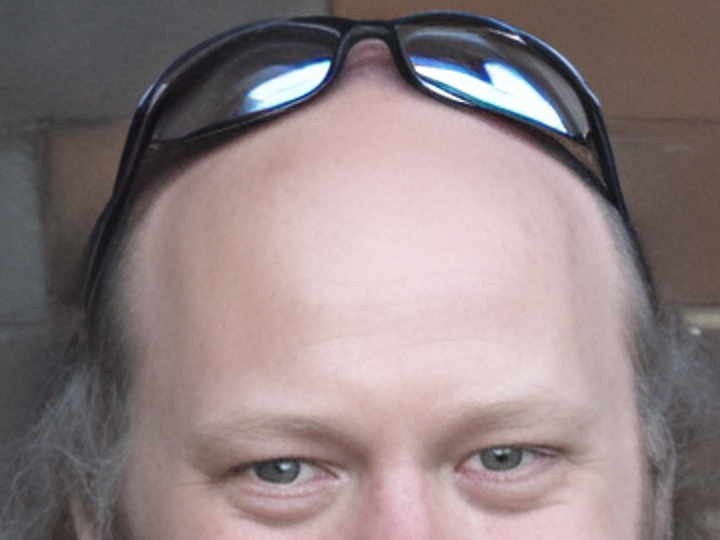} \hspace{-4.5mm} 
				\\
				Low-Quality Input \hspace{-2mm} &
				Real-ESRGAN+\cite{wang2021real} \hspace{-3.5mm} &
				StableSR\cite{wang2023exploiting} \hspace{-3.5mm}
				\\
				\includegraphics[width=0.19\textwidth]{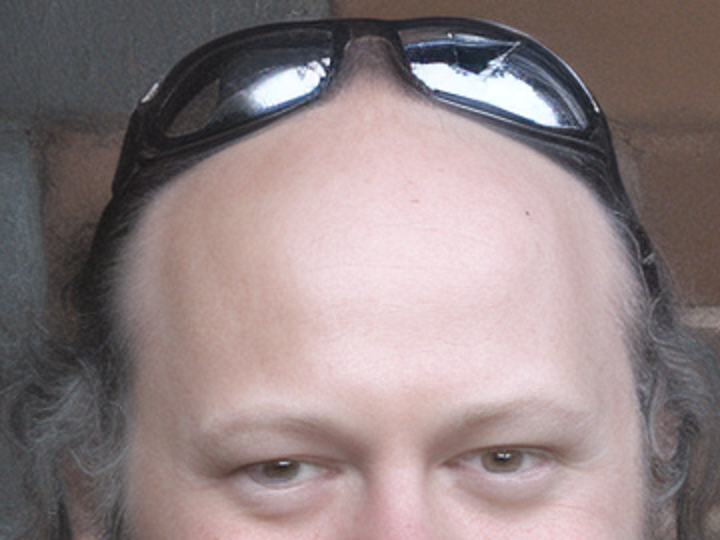} \hspace{-4.5mm} &
				\includegraphics[width=0.19\textwidth]{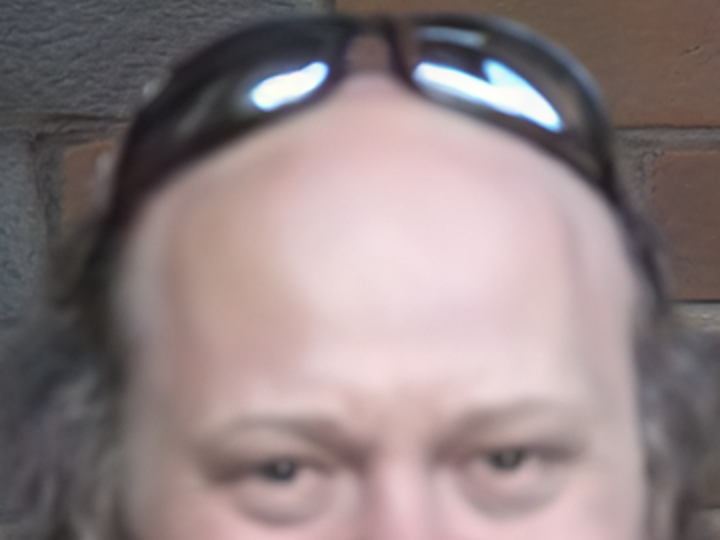} \hspace{-4.5mm} &
				\includegraphics[width=0.19\textwidth]{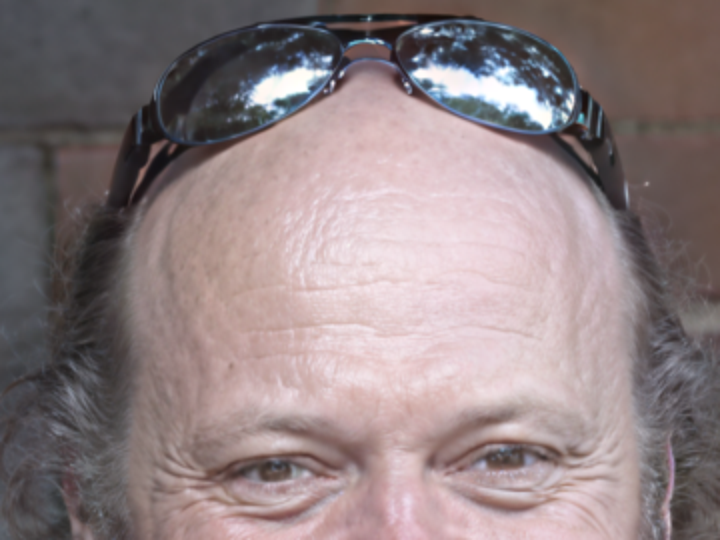} \hspace{-4.5mm}
				\\ 
				DiffBIR\cite{lin2023diffbir} \hspace{-3.5mm} &
				PASD\cite{yang2023pixel} \hspace{-3.5mm} &
				\method{} (ours) \hspace{-3.5mm}
			\end{tabular}
		\end{adjustbox}
		\hspace{-2mm}
        \\
        		\hspace{-0.42cm}
		\begin{adjustbox}{valign=t}
			\begin{tabular}{c}
				\includegraphics[width=0.409\textwidth]{{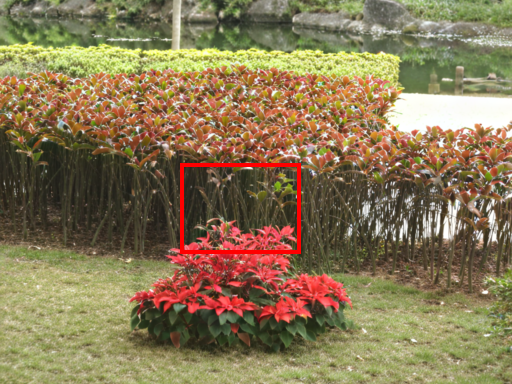}}
                \\
                \method{} (ours)
				\hspace{-10mm}
			\end{tabular}
		\end{adjustbox}
		\hspace{-5mm}
		\begin{adjustbox}{valign=t}
			\begin{tabular}{ccc}
				\includegraphics[width=0.19\textwidth]{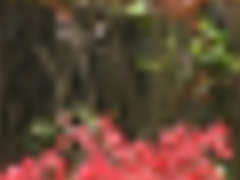} \hspace{-4.5mm} &
				\includegraphics[width=0.19\textwidth]{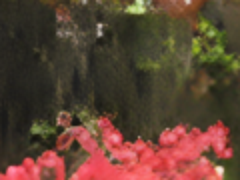} \hspace{-4.5mm} &
				\includegraphics[width=0.19\textwidth]{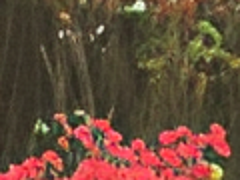} \hspace{-4.5mm} 
				\\
				Low-Quality Input \hspace{-2mm} &
				Real-ESRGAN+\cite{wang2021real} \hspace{-3.5mm} &
				StableSR\cite{wang2023exploiting} \hspace{-3.5mm}
				\\
				\includegraphics[width=0.19\textwidth]{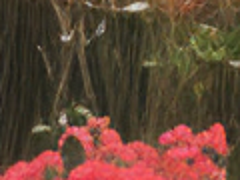} \hspace{-4.5mm} &
				\includegraphics[width=0.19\textwidth]{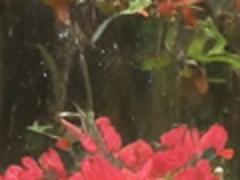} \hspace{-4.5mm} &
				\includegraphics[width=0.19\textwidth]{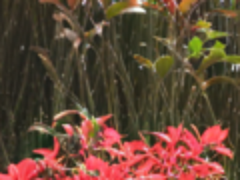} \hspace{-4.5mm}
				\\ 
				DiffBIR\cite{lin2023diffbir} \hspace{-3.5mm} &
				PASD\cite{yang2023pixel} \hspace{-3.5mm} &
				\method{} (ours) \hspace{-3.5mm}
			\end{tabular}
		\end{adjustbox}
		\hspace{-2mm}
        \\
        		\hspace{-0.42cm}
		\begin{adjustbox}{valign=t}
			\begin{tabular}{c}
				\includegraphics[width=0.409\textwidth]{{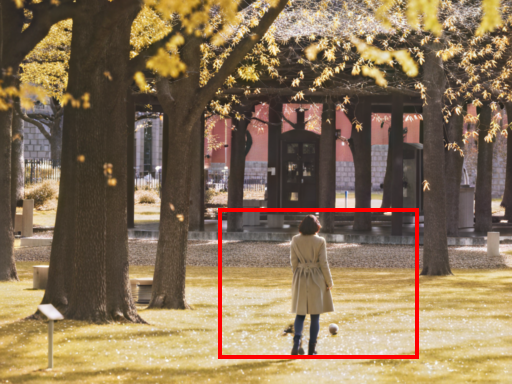}}
                \\
                \method{} (ours)
				\hspace{-10mm}
			\end{tabular}
		\end{adjustbox}
		\hspace{-5mm}
		\begin{adjustbox}{valign=t}
			\begin{tabular}{ccc}
				\includegraphics[width=0.19\textwidth]{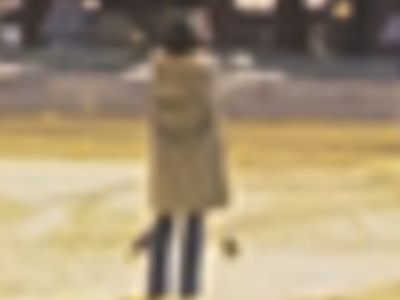} \hspace{-4.5mm} &
				\includegraphics[width=0.19\textwidth]{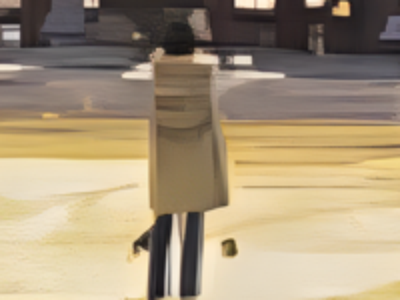} \hspace{-4.5mm} &
				\includegraphics[width=0.19\textwidth]{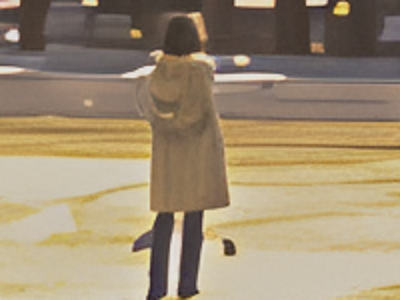} \hspace{-4.5mm} 
				\\
				Low-Quality Input \hspace{-2mm} &
				Real-ESRGAN+\cite{wang2021real} \hspace{-3.5mm} &
				StableSR\cite{wang2023exploiting} \hspace{-3.5mm}
				\\
				\includegraphics[width=0.19\textwidth]{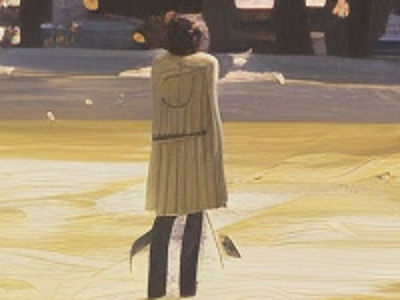} \hspace{-4.5mm} &
				\includegraphics[width=0.19\textwidth]{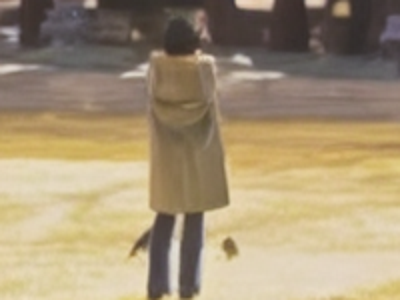} \hspace{-4.5mm} &
				\includegraphics[width=0.19\textwidth]{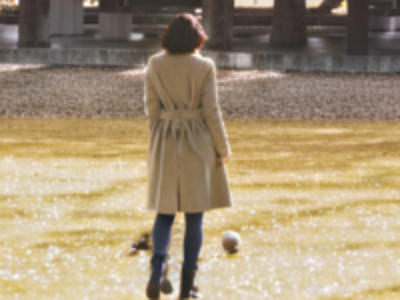} \hspace{-4.5mm}
				\\ 
				DiffBIR\cite{lin2023diffbir} \hspace{-3.5mm} &
				PASD\cite{yang2023pixel} \hspace{-3.5mm} &
				\method{} (ours) \hspace{-3.5mm}
			\end{tabular}
		\end{adjustbox}
	\end{tabular}
    }
    % \hspace{-8mm}
    \vspace{-4mm}
	\caption{Qualitative comparison with different methods. Our method can accurately restore the texture and details of the corresponding object under challenging degradation. Zoom in for better view.}
    \label{fig:supp_visual2}
	\vspace{-6mm}
\end{figure*}
% %%%%%%%%%%%%%%%%%%%%%%%%%%%%%%%%%%%%%%%%%%%%%%%%%%

%%%%%%%%%%%%%%%%%%%%%%%%%%%%%%%%%%%%%%%%%%%%%%%%%%

\begin{figure*}[t]
	%\newlength-4mm
	%\setlength{-4mm}{-0.4cm}
	\scriptsize
	\centering
    \includegraphics[width=\linewidth]{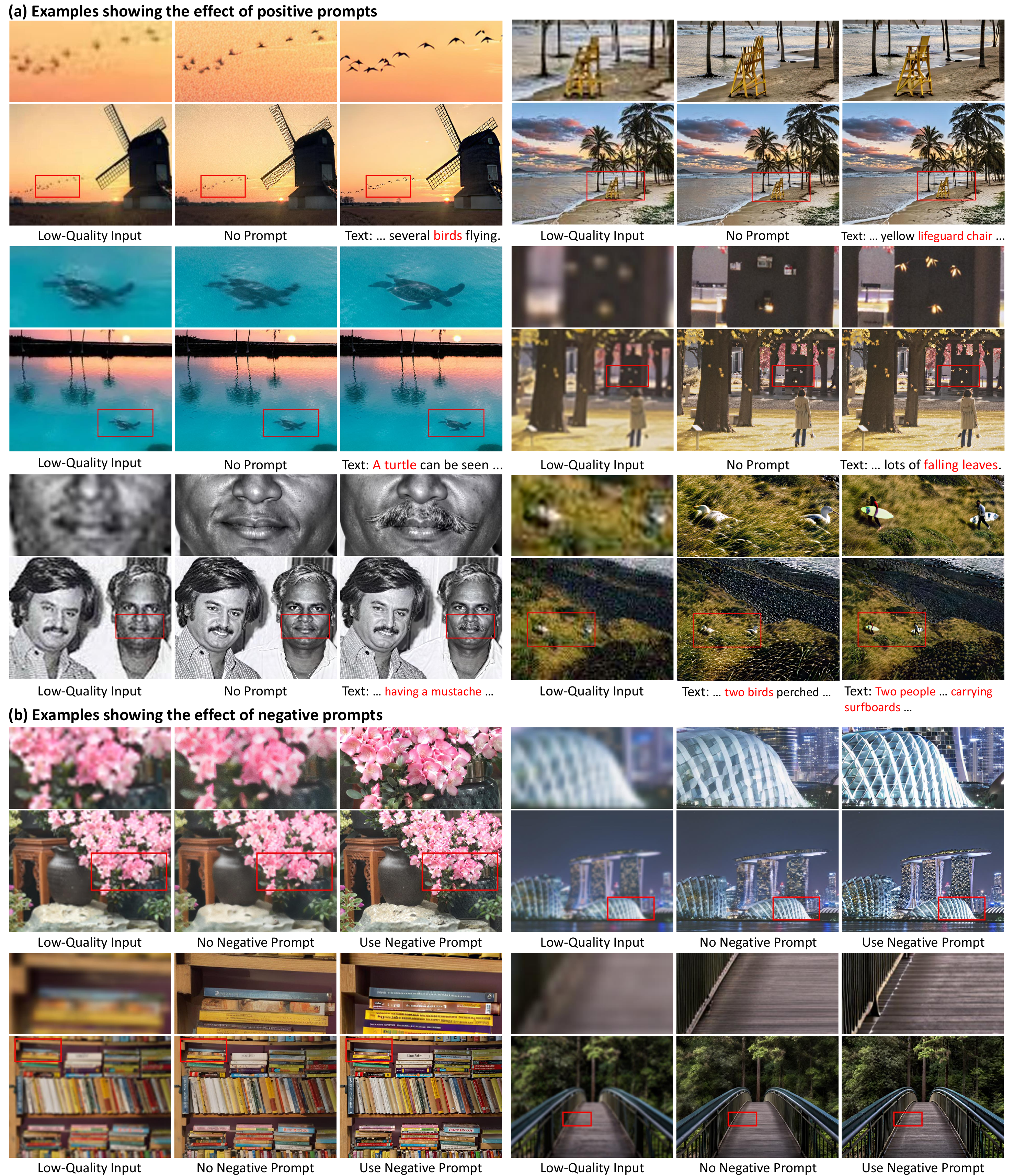}
	\vspace{-6mm}
	\caption{More visual results of the text prompts' influences. (a) and (b) show the examples of positive prompts and negative prompts, respectively. Zoom in for better view.}
    \label{fig:supp_prompt}
	\vspace{-5mm}
\end{figure*}

%%%%%%%%%%%%%%%%%%%%%%%%%%%%%%%%%%%%%%%%%%%%%%%%%%

% \input{sec/supp}

\end{document}